\documentclass{article} 
\usepackage{iclr2025_conference,times}

\usepackage{amsmath,amsfonts,bm}

\def\eqref#1{equation~\ref{#1}}

\def\1{\bm{1}}

\DeclareMathAlphabet{\mathsfit}{\encodingdefault}{\sfdefault}{m}{sl}
\SetMathAlphabet{\mathsfit}{bold}{\encodingdefault}{\sfdefault}{bx}{n}

\usepackage[hidelinks]{hyperref}
\usepackage{url}
\usepackage{booktabs}       
\usepackage{amsfonts}       
\usepackage{nicefrac}       
\usepackage{microtype}      
\usepackage{xcolor}         
\usepackage{wrapfig}

\usepackage{times}
\usepackage{latexsym}
\usepackage{booktabs}
\usepackage{makecell}
\usepackage{multirow}
\usepackage{graphicx}
\usepackage{amsmath}
\usepackage{amsfonts}
\usepackage{xcolor}
\usepackage{todonotes}
\usepackage{listings}
\usepackage{subcaption}
\usepackage{wrapfig}
\usepackage{floatrow}

\usepackage{colortbl}

\usepackage{amsmath}
\usepackage{amssymb}
\usepackage{bm}

\usepackage[bottom]{footmisc}

\usepackage{tcolorbox}
\usepackage[linesnumbered,ruled,vlined]{algorithm2e}

\title{LongLLaDA: Unlocking Long Context \\ Capabilities in Diffusion LLMs}

\author{%
Xiaoran Liu\textsuperscript{1,2}, Yuerong Song\textsuperscript{1,2}, Zhigeng Liu\textsuperscript{1,2}, \\
\textbf{Zengfeng Huang\textsuperscript{1,2}, Qipeng Guo\textsuperscript{2,3}, Ziwei He\textsuperscript{2}\thanks{\ \ Corresponding Author.}\ , 
Xipeng Qiu\textsuperscript{1,2}\footnotemark[1]} \\
\textsuperscript{1}Fudan University, \textsuperscript{2}Shanghai Innovation Institute,  \textsuperscript{3}Shanghai AI Lab \\
\texttt{xrliu24@m.fudan.edu.cn},\;
\texttt{ziwei.he@sii.edu.cn},\;
\texttt{xpqiu@fudan.edu.cn}\\
}

\usepackage{tabu}

\iclrfinalcopy 

\begin{document}

\maketitle

\begin{abstract}

Large Language Diffusion Models, or diffusion LLMs, have emerged as a significant focus in NLP research, with substantial effort directed toward understanding their scalability and downstream task performance. However, their long-context capabilities remain unexplored, lacking systematic analysis or methods for context extension. In this work, we present the first systematic investigation comparing the long-context performance of diffusion LLMs and traditional auto-regressive LLMs.
We first identify a unique characteristic of diffusion LLMs, unlike auto-regressive LLMs, they maintain remarkably \textbf{\textit{stable perplexity}} during direct context extrapolation. Moreover, where auto-regressive models fail outright during the Needle-In-A-Haystack task with context exceeding their pretrained length, we discover diffusion LLMs exhibit a distinct ``\textbf{\textit{local perception}}" phenomenon, enabling successful retrieval from recent context segments. We explain both phenomena through the lens of Rotary Position Embedding (RoPE) scaling theory. Building on these observations, we propose LongLLaDA, a training-free method that integrates LLaDA with the NTK-based RoPE extrapolation. Our results validate that established extrapolation scaling laws remain effective for extending the context windows of diffusion LLMs. Furthermore, we identify long-context tasks where diffusion LLMs outperform auto-regressive LLMs and others where they fall short.
Consequently, this study establishes the first length extrapolation method for diffusion LLMs while providing essential theoretical insights and empirical benchmarks critical for advancing future research on long-context diffusion LLMs. The code is available at \url{https://github.com/OpenMOSS/LongLLaDA}.

\end{abstract}

\section{Introduction}\label{intro}

Recently, diffusion LLMs have become widely discussed in Natural Language Processing research~\citep{nie2025large,dream2025}. They are regarded as a potential solution to key limitations of traditional auto-regressive LLMs~\citep{touvron2023llama,Sun2024MOSS}, including the reversal curse~\citep{berglund2023reversal}, complex reasoning~\citep{dziri2023faith}, and maintaining coherence across extended contexts~\citep{bachmann2024pitfalls,ye2024beyond,dream2025}. Significant research efforts have focused on validating their scalability~\citep{nie2025large,dream2025}, adapting them for multimodality~\citep{yang2025mmada,you2025llada,yu2025dimple}, applying them to reasoning tasks~\citep{zhao2025d1,huang2025reinforcing,zhu2025llada}, and optimizing their efficiency~\citep{ma2025dkv,hu2025accelerating,wu2025fast}. However, the long-context capabilities of diffusion LLMs, specifically their performance and potential for length extrapolation, remain unexplored.

\begin{figure}[!t]
\begin{minipage}{0.98\textwidth}
    \begin{subfigure}[b]{0.48\linewidth}
        \centering
        \includegraphics[width=\linewidth]{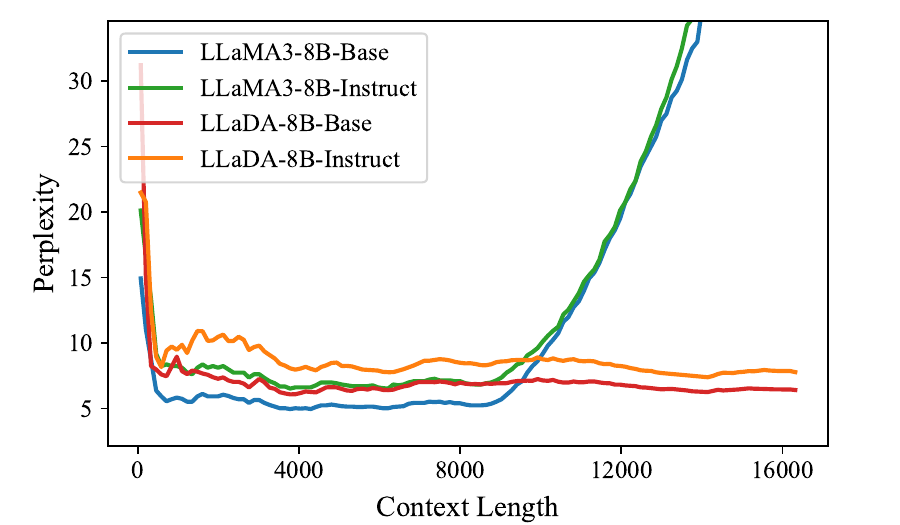}
        \caption{Comparison of Perplexity}
        \label{diffusion_ppl_gov}
    \end{subfigure}
    \hfill
    \begin{subfigure}[b]{0.48\linewidth}
        \centering
        \includegraphics[width=\linewidth]{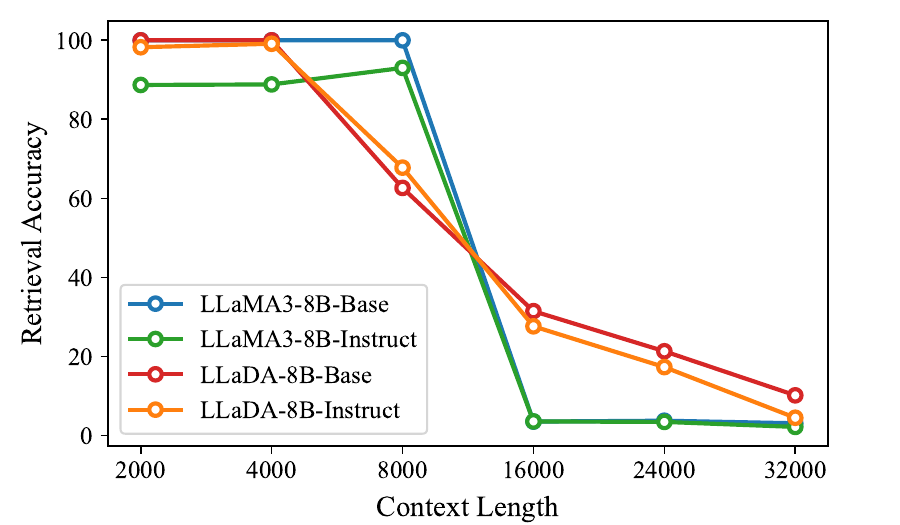}
        \caption{Comparison of Retrieval Accuracy}
        \label{diffusion_niah}
    \end{subfigure}
    \caption{Comparison of perplexity and retrieval accuracy between the diffusion LLM, LLaDA-8B, and the auto-regressive LLM, LLaMA3-8B, both within and beyond pre-training context length.\label{fig_intro}}
\end{minipage}
\end{figure}

We begin by systematically evaluating diffusion LLM LLaDA~\citep{nie2025large} against auto-regressive LLM LLaMA3~\citep{meta2024introducing} on perplexity and retrieval tasks, both within and beyond their pretrained context lengths (Figure~\ref{fig_intro}). Notably, diffusion LLMs maintain stable perplexity and exhibit localized perception during direct length extrapolation. In stark contrast, auto-regressive LLMs suffer catastrophic perplexity surges and performance collapse when input length exceeds their maximum supported context window, 8k tokens. This divergence reveals fundamental architectural differences in long-context handling, raising critical questions: (1) What mechanisms enable diffusion LLMs’ extrapolation stability? (2) Can established length-extension techniques for auto-regressive LLMs be transferred to diffusion architectures? (3) How do diffusion LLMs perform on long-context benchmarks relative to auto-regressive baselines, and what unique capabilities or limitations emerge?

In this work, we address these questions through comprehensive experiments and analysis. Besides the perplexity and retrieval experiment, we also benchmark Needle-In-A-Haystack (NIAH) performance for diffusion LLMs (LLaDA~\citep{nie2025large}, LLaDA-1.5~\citep{zhu2025llada}, Dream-v0~\citep{dream2025}), quantitatively confirming their local perception bias during length extrapolation. We then analyze this phenomenon through Rotary Position Embedding (RoPE) theory, validating our interpretation with t-SNE visualizations. Building on these insights, we propose LongLLaDA, a training-free method which successfully extends LLaDA's context window using NTK-based RoPE extrapolation~\citep{fixedNTK}, and verify preserved scaling laws~\citep{liu2023scaling}. Finally, we identify task-dependent capabilities where diffusion LLMs surpass or lag behind auto-regressive counterparts on long-context benchmarks. Our contributions are summarized as follows:

\begin{itemize}
    \itemsep0.3mm
    \item \textbf{First systematic analysis} of diffusion LLMs' long-context behavior, revealing their unique characteristics for stable perplexity and local perception during context extrapolation, with mechanistic explanation via RoPE dynamics.
    \item \textbf{Effective context extension} demonstrating NTK-based RoPE extrapolation and scaling laws transfer seamlessly to diffusion LLMs, achieving 6$\times$ context expansion (24k tokens) without further training.
    \item \textbf{Capability benchmarking} revealing diffusion LLMs match auto-regressive models on retrieval tasks, lag in aggregation, but excel at synthetic QA. We provide foundational insights for future long-context diffusion research.
\end{itemize}

\section{Long-Context Phenomenology of Diffusion LLMs}\label{observation}

\begin{figure}[!tb]
\begin{minipage}{0.98\textwidth}
    \begin{subfigure}[b]{0.48\linewidth}
        \centering
        \includegraphics[width=\linewidth]{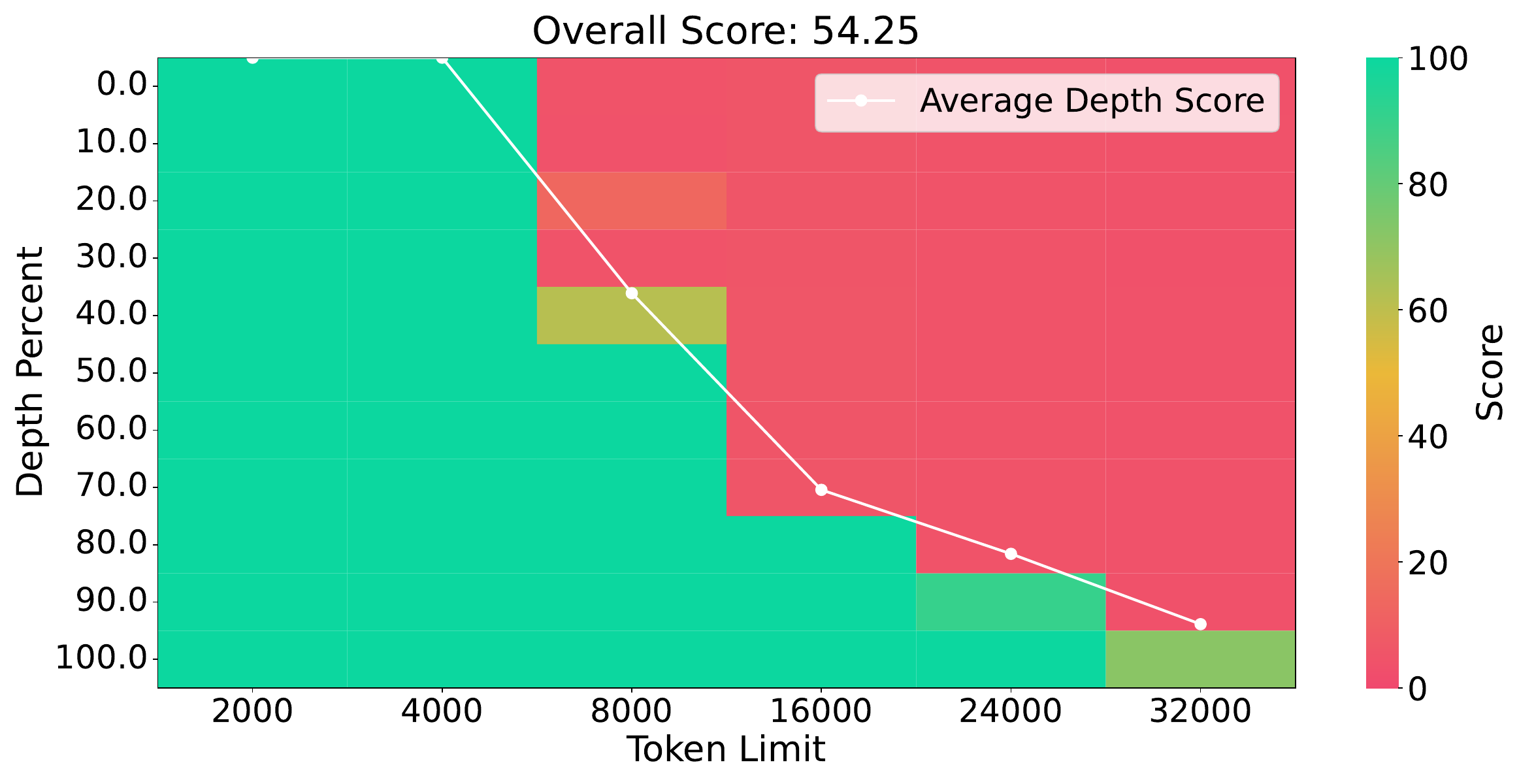}
        \caption{LLaDA-8B-Base}
        \label{llada_8b_base_niah}
    \end{subfigure}
    \hfill
    \begin{subfigure}[b]{0.48\linewidth}
        \centering
        \includegraphics[width=\linewidth]{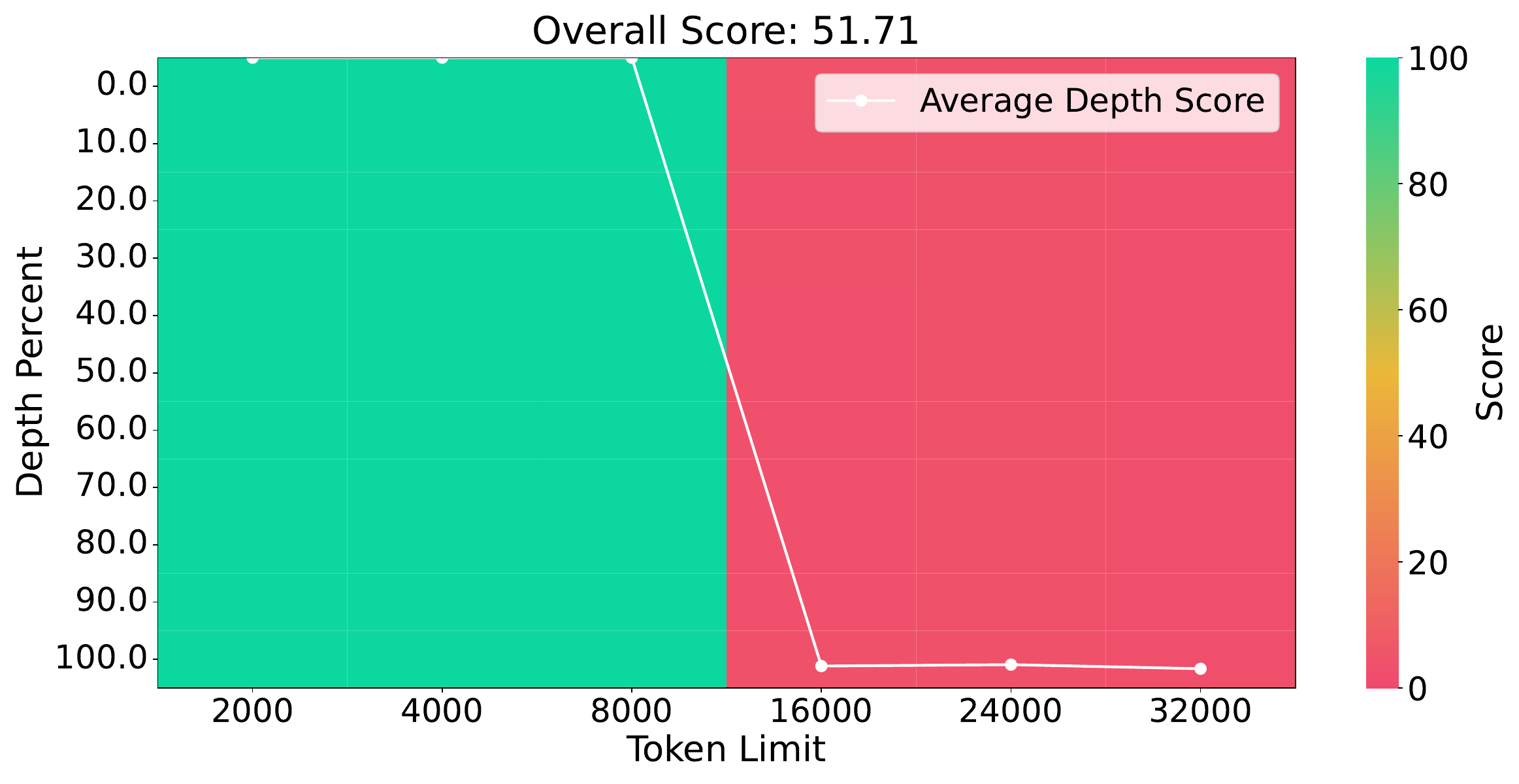}
        \caption{LLaMA3-8B-Base}
        \label{llama3_8b_base_niah}
    \end{subfigure}
    \vskip\baselineskip
    \begin{subfigure}[b]{0.48\linewidth}
        \centering
        \includegraphics[width=\linewidth]{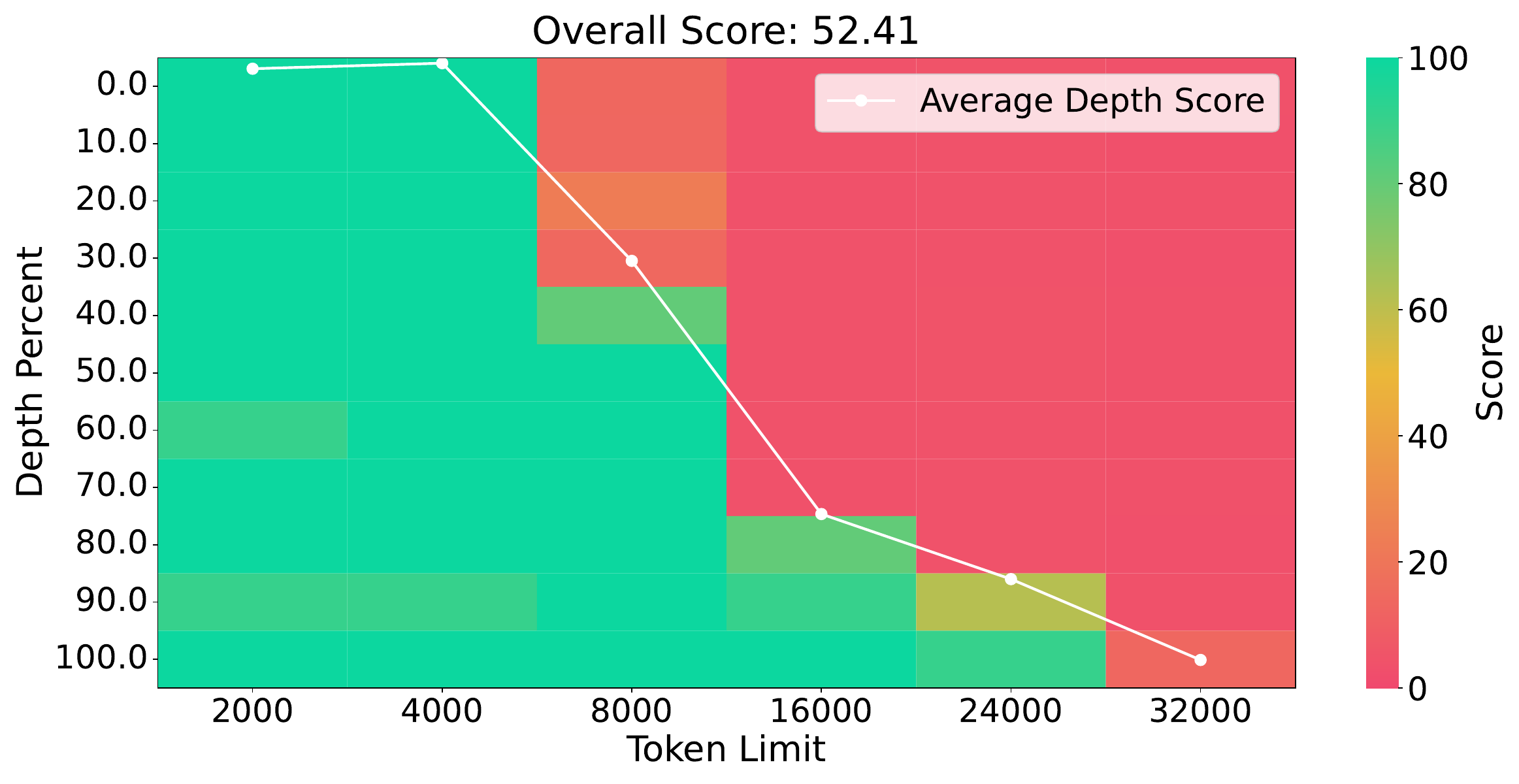}
        \caption{LLaDA-8B-Instruct}
        \label{llada_8b_chat_niah}
    \end{subfigure}
    \hfill
    \begin{subfigure}[b]{0.48\linewidth}
        \centering
        \includegraphics[width=\linewidth]{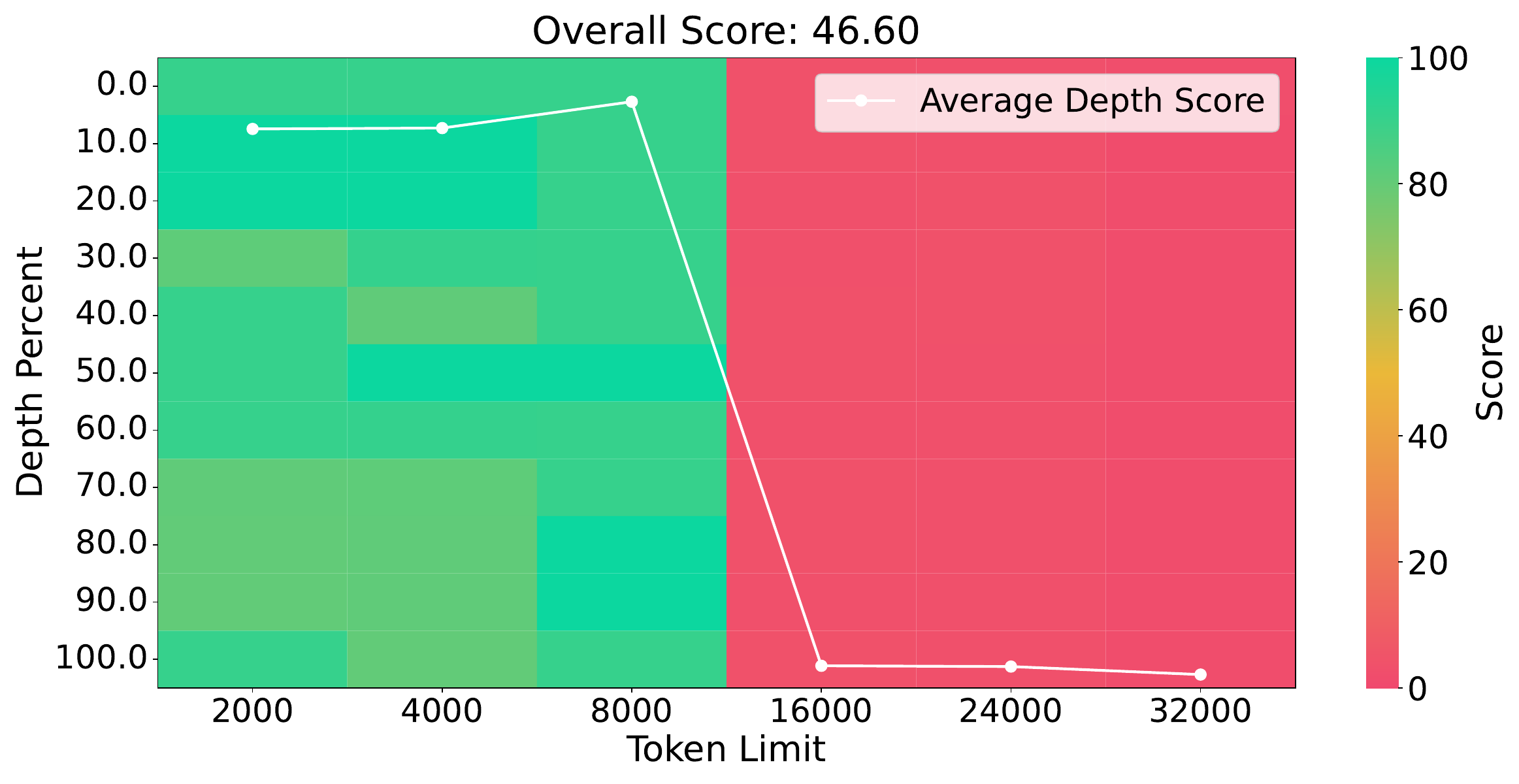}
        \caption{LLaMA3-8B-Instruct}
        \label{llama3_8b_chat_niah}
    \end{subfigure}
    \caption{Results of Needle-In-A-Haystack tests~\citep{needle_in_a_haystack} on LLaDA-8B Series~\citep{nie2025large} and LLaMA3-8B Series~\citep{meta2024llama} under direct extrapolation.\label{fig_direct}}
\end{minipage}
\end{figure}

\begin{figure}[!tb]
\begin{minipage}{0.98\textwidth}
    \begin{subfigure}[b]{0.48\linewidth}
        \centering
        \includegraphics[width=\linewidth]{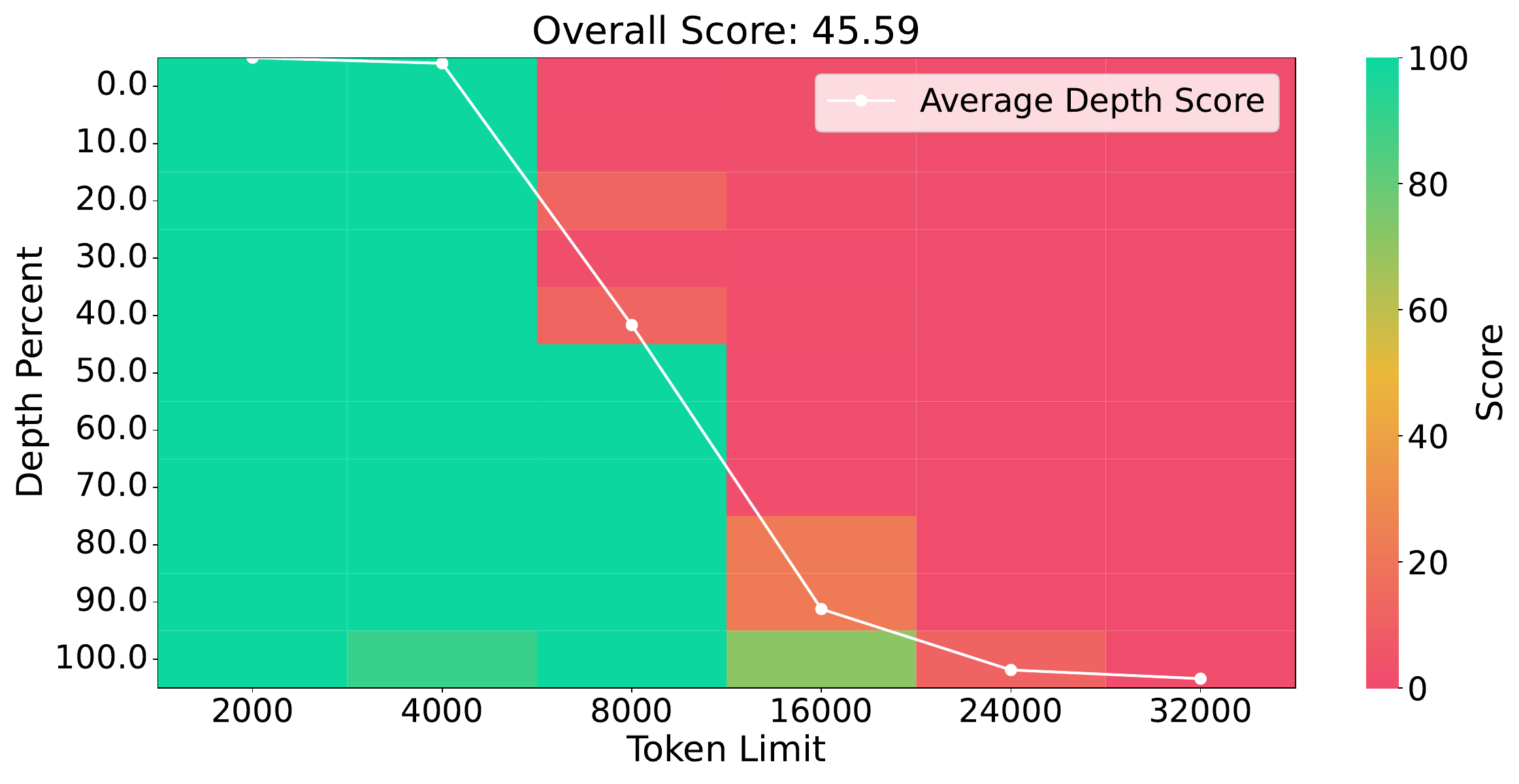}
        \caption{LLaDA-8B-Base with $s=1$}
        \label{llada_8b_base_niah_step1}
    \end{subfigure}
    \hfill
    \begin{subfigure}[b]{0.48\linewidth}
        \centering
        \includegraphics[width=\linewidth]{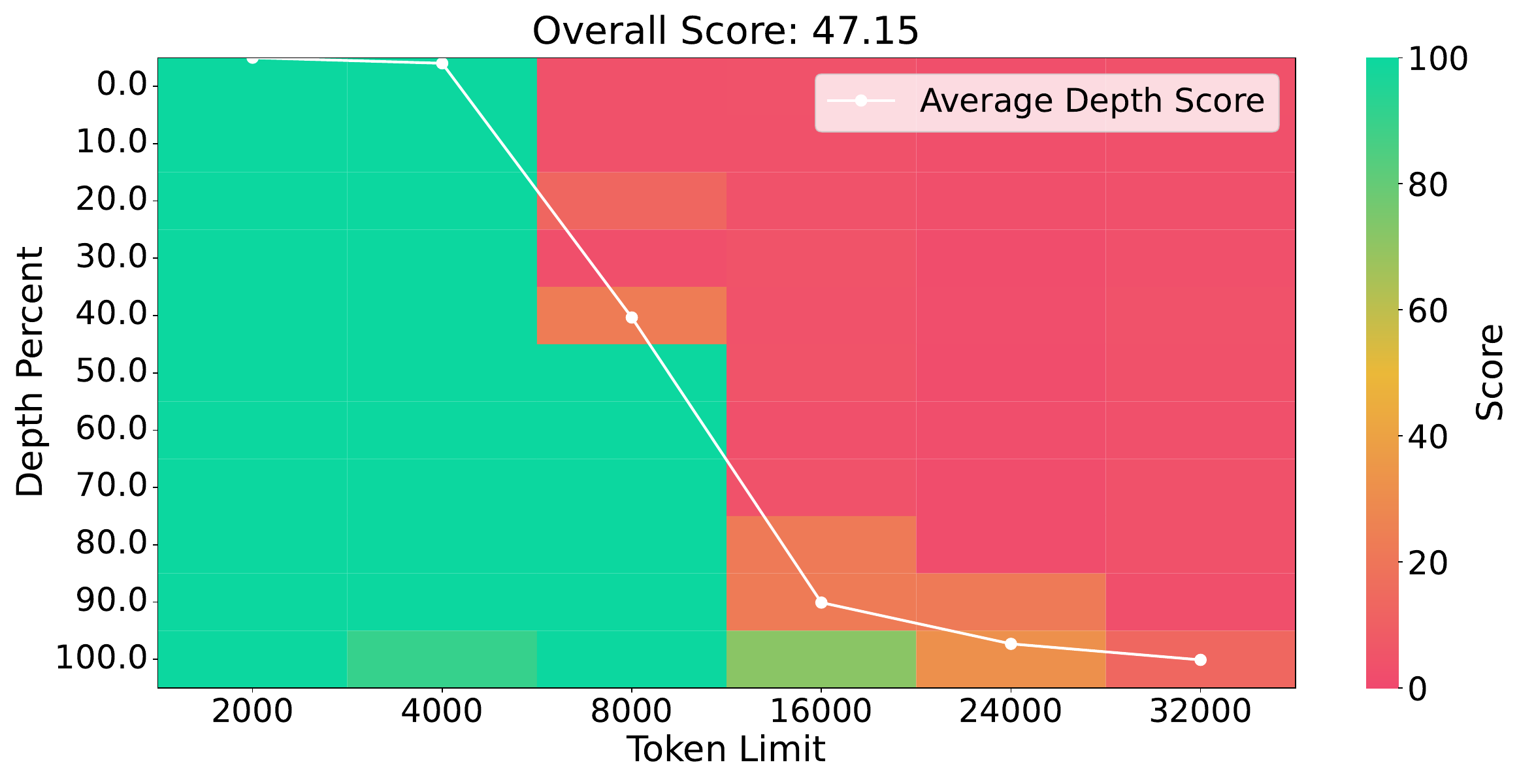}
        \caption{LLaDA-8B-Base with $s=4$}
        \label{llada_8b_base_niah_step4}
    \end{subfigure}
    \vskip\baselineskip
    \begin{subfigure}[b]{0.48\linewidth}
        \centering
        \includegraphics[width=\linewidth]{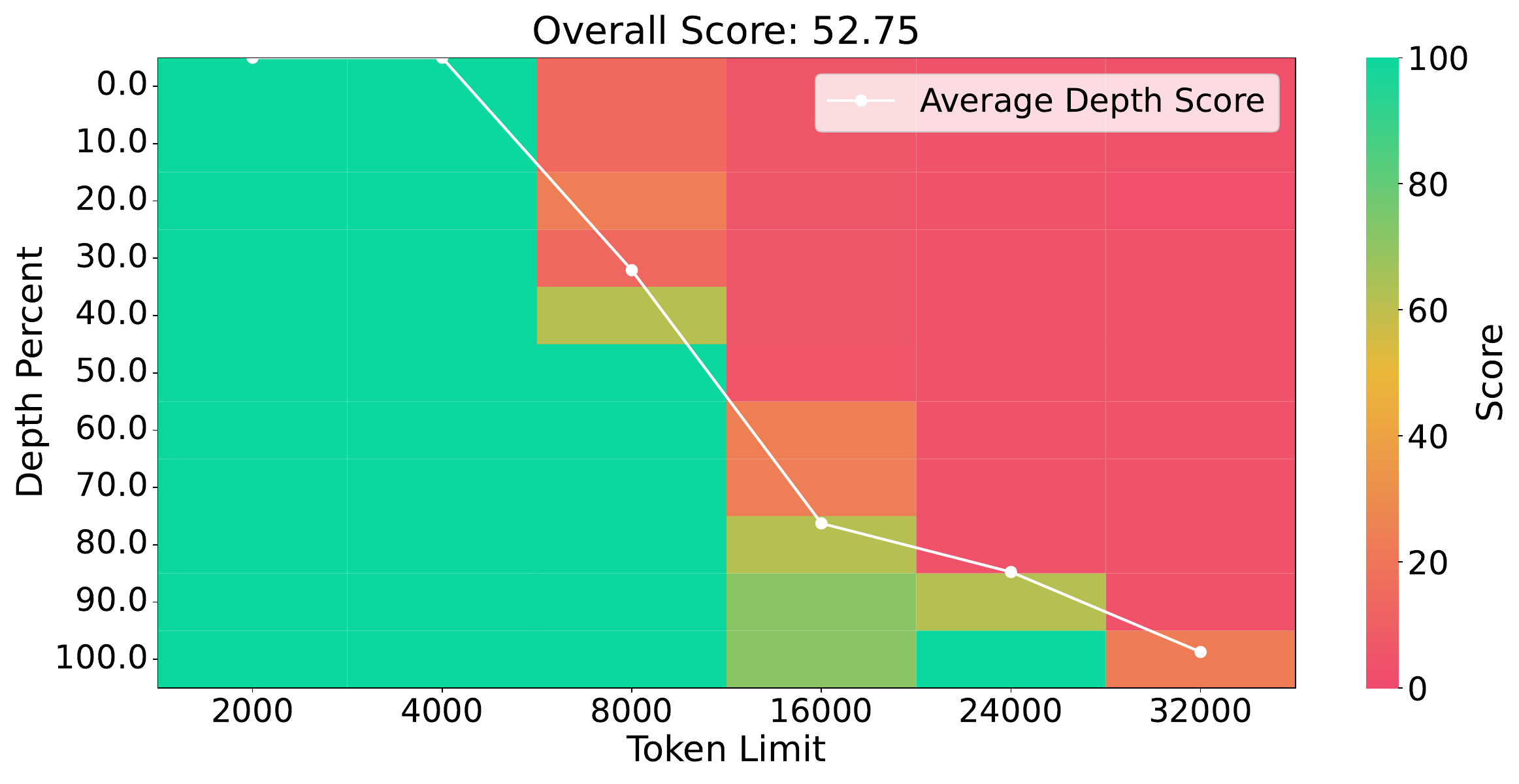}
        \caption{LLaDA-8B-Base with $s=8$}
        \label{llada_8b_base_niah_step8}
    \end{subfigure}
    \hfill
    \begin{subfigure}[b]{0.48\linewidth}
        \centering
        \includegraphics[width=\linewidth]{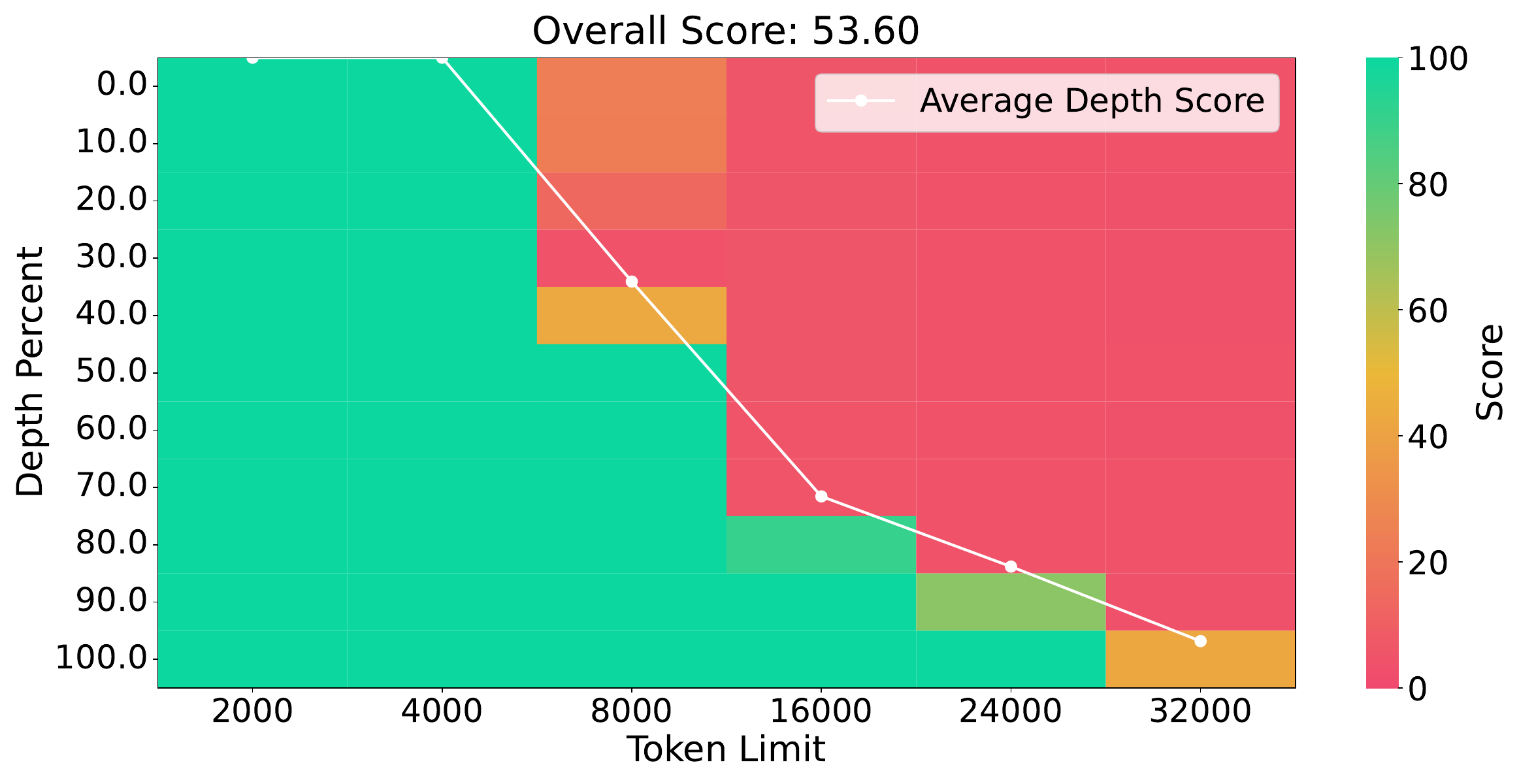}
        \caption{LLaDA-8B-Base with $s=16$}
        \label{llada_8b_base_niah_step16}
    \end{subfigure}
    \caption{NIAH Results of LLaDA-8B-Base~\citep{nie2025large} with different sampling steps $s$.\label{llada_base_step}}
\end{minipage}
\end{figure}

We first evaluate the length extrapolation capabilities of diffusion LLMs, including LLaDA~\citep{nie2025large}, LLaDA-1.5~\citep{zhu2025llada}, and Dream-v0~\citep{dream2025}, compared with auto-regressive LLMs such as LLaMA3~\citep{meta2024introducing}, via Needle-In-A-Haystack~\citep{needle_in_a_haystack,li2024needlebench}, based on the experimental setup in Appendix~\ref{setup}. All LLMs are required to generate at most 32 tokens, with diffusion LLMs using a block size and sampling steps of 32. The results are shown in Figure~\ref{fig_direct}. LLaMA3-8B-Base and LLaMA3-8B-Instruct maintain perfect retrieval accuracy within their pretrained 8k length, but suffer catastrophic performance degradation beyond this limit, failing to retrieve information at any depth. In contrast, LLaDA-8B-Base and LLaDA-8B-Instruct achieve 100\% retrieval accuracy within a 4k context. Surprisingly, when exceeding 4k, up to 24k, LLaDA still retrieves information from the nearest 4k window, demonstrating a local perception like a sliding window. This behavior remarkably differs from auto-regressive LLM extrapolation. Similar phenomena are observed in LLaDA-1.5 and Dream-v0, as illustrated in Appendix~\ref{more_results}.

Different from auto-regressive LLMs, diffusion LLMs are influenced by sampling steps and strategies. For simplicity, we compare the impact of sampling steps on retrieval depth in NIAH. As shown in Figure~\ref{llada_base_step}, using the same input-output settings from previous experiments, we evaluate the LLaDA-8B-Base with sampling step $s=1,4,8,16$. Results show that at 1 or 4 steps, LLaDA-8B-Base fails to retrieve information beyond 8k length, and increasing $s$ to 8 or 16 can achieve retrieval depths of 25\% at 16k and almost 10\% at 24k context length. Similar results are observed on LLaDA-8B-Instruct and LLaDA-1.5 in Appendix~\ref{more_results}, demonstrating that the long-context performance of diffusion LLMs is influenced by sampling steps, but remains constrained by the maximum supported context length.

\section{Mechanistic Analysis}\label{obs_direct}

\begin{figure}[!tb]
\begin{minipage}{0.98\textwidth}
    \begin{subfigure}[b]{0.48\linewidth}
        \centering
        \includegraphics[width=\linewidth]{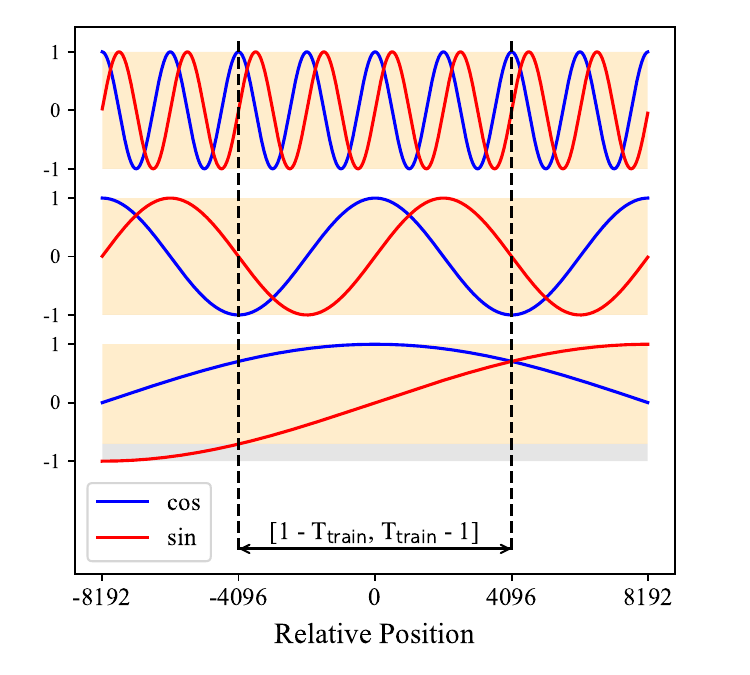}
        \caption{LLaDA with $T_\text{train}=$ 4k}
        \label{rope_auto_regressive}
    \end{subfigure}
    \hfill
    \begin{subfigure}[b]{0.48\linewidth}
        \centering
        \includegraphics[width=\linewidth]{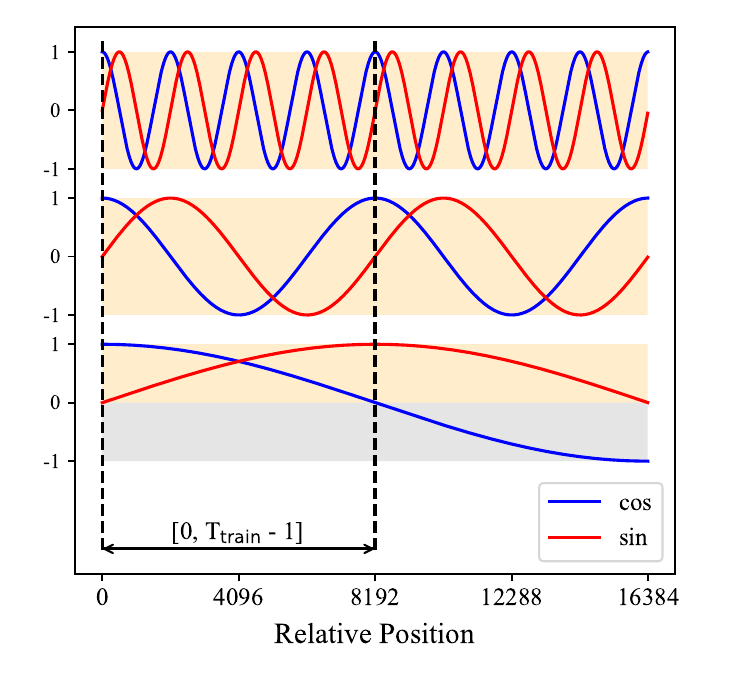}
        \caption{LLaMA3 with $T_\text{train}=$ 8k}
        \label{rope_diffusion}
    \end{subfigure}
    \caption{Comparison of trained position embedding interval between LLaDA-8B and LlaMA3-8B. The area within the dashed line represents trained relative position, while that beyond represents the relative position in length extrapolation, with unlearned position embedding values colored in gray. \label{sinusoid}}
\end{minipage}
\end{figure}

According to the preliminary knowledge in Appendix~\ref{preliminary}, we attribute this phenomenon to diffusion LLMs being trained with richer positional information compared to auto-regressive LLMs. Critically, the bidirectional attention mechanisms in diffusion LLMs expose them to relative position rage of $[1-T_\text{train}, T_\text{train}-1]$ during training, contrasting with the $[0, T_\text{train}-1]$ range typical of auto-regressive models. This difference is evident in the RoPE mechanism. As visualized in Figure~\ref{sinusoid}, for LLaDA ($T_\text{train}=$ 4k) and LLaMA ($T_\text{train}=$ 8k), we observe how the positional embeddings (sine/cosine components) behave within and beyond their maximum trained relative positions. 
\begin{itemize}
    \item \textbf{High Frequencies}: Both models perceive complete sinusoidal periods within their maximum trained relative distance, yielding comparable positional information encoding.
    \item \textbf{Moderate Frequencies}: LLaMA3's auto-regressive attention observes relative positions $[0, 8191]$ when trained on 8192-token sequences. In contrast, LLaDA's bidirectional attention observes symmetric relative positions $[-4095, 4095]$ despite its shorter 4096-token training length. This symmetric coverage provides a key advantage by fully capturing a complete period of both the cosine and sine, enhancing its tolerance of direct length extrapolation.
    \item \textbf{Low Frequencies}: Both models exhibit limited extrapolation capability beyond their pretrained context windows. However, as visualized in Figure~\ref{sinusoid}, the out-of-distribution (OOD) regions differ remarkably: LLaMA3 struggles to capture all negative position embeddings (gray region), representing half of the potential embedding space, while LLaDA significantly reduces the unlearned OOD spaces, resulting in enhanced robustness in length extrapolation.
\end{itemize}

This results in a relatively flattened perplexity growth curve, similar to auto-regressive RoPE-based LLMs with a smaller base~\citep{liu2023scaling,men2024base}, as detailed in Appendix~\ref{preliminary}. However, since the cosine function in RoPE, which primarily captures relative distances, is even, negative relative positions do not increase the LLM's maximum perceivable distance in the pre-training stage. Thus, diffusion LLM can only retrieve key information from limited relative positions within the training length, leading to the observed decay pattern in the NIAH evaluation. 

\begin{figure}[!tb]
\begin{minipage}{0.98\textwidth}
    \begin{subfigure}[b]{0.48\linewidth}
        \centering
        \includegraphics[width=\linewidth]{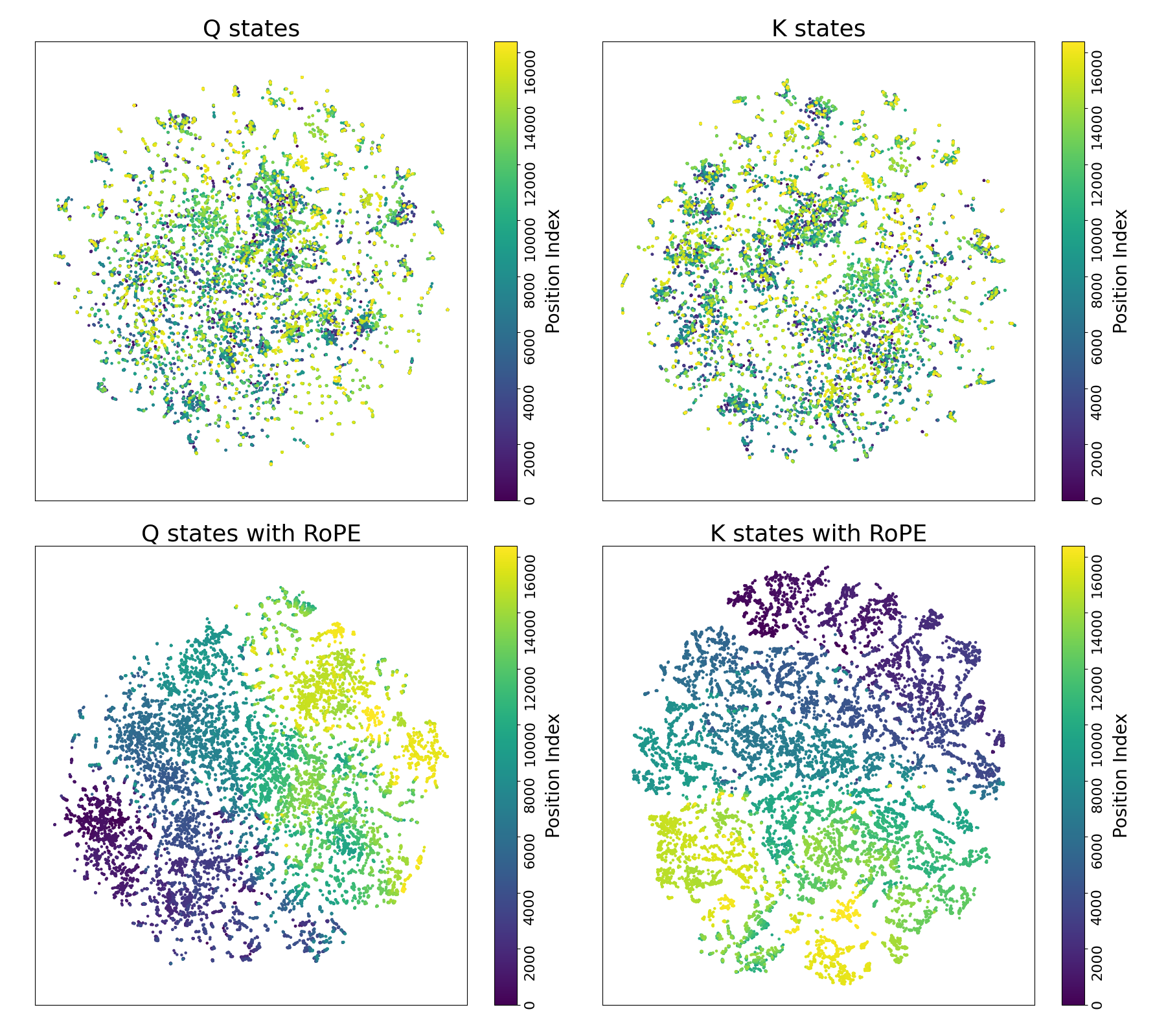}
        \caption{LLaDA-8B-Base}
        \label{tsne_auto_regressive}
    \end{subfigure}
    \hfill
    \begin{subfigure}[b]{0.48\linewidth}
        \centering
        \includegraphics[width=\linewidth]{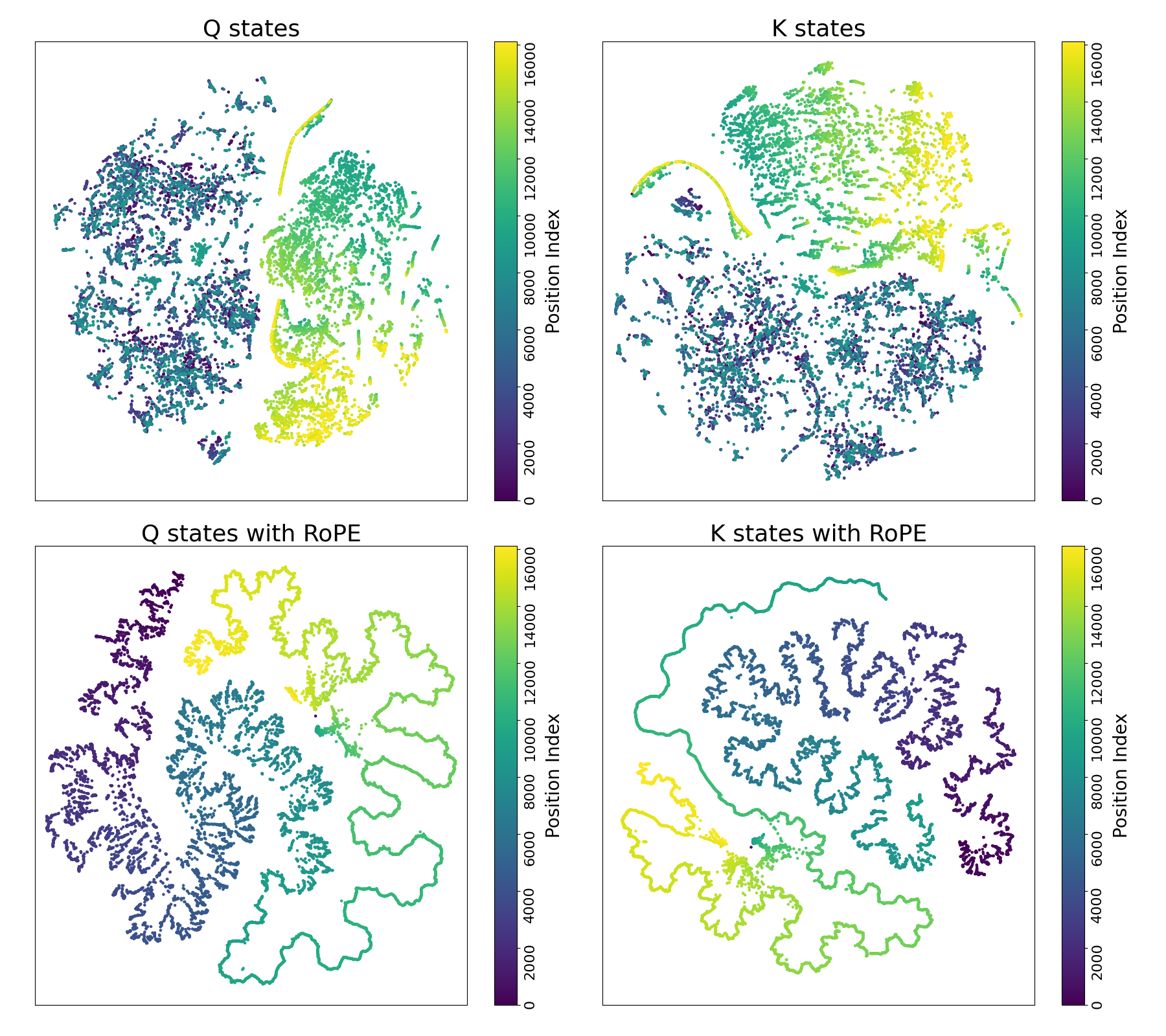}
        \caption{LLaMA3-8B-Base}
        \label{tsne_diffusion}
    \end{subfigure}
    \caption{Visualization of the QK states from the final layer of LLaMA3-8B-Base~\citep{meta2024introducing} and LLaDA-8B-Base~\citep{nie2025large} for sample from the GovReport subsets in LongBench~\citep{bai2023longbench}. The visualization uses a 2D t-SNE projection~\citep{van2008visualizing}, with each token represented as a point in the image and the position index shown via color changing.\label{tsne}}
\end{minipage}
\end{figure}

We validate this interpretation with the t-SNE visualization~\citep{van2008visualizing,zandieh2024subgen} of QK states from the final layer of LLaMA3-8B-Base~\citep{meta2024introducing} and LLaDA-8B-Base~\citep{nie2025large}, as shown in Figure~\ref{tsne}. As shown in Figure~\ref{tsne_auto_regressive}, for auto-regressive LLMs such as LLaMA3-8B-Base, the QK states within and beyond the maximum supported context length, 8k, present two different distribution clusters, and the manifold for QK states with RoPE also shows a different trend when position embedding becomes OOD. Comparatively, regarding the clustering feature for diffusion LLMs such as LLaDA-8B-Base, there is no distribution shift between QK states within and beyond 4k, and a uniform manifold for QK states with RoPE. This demonstrates that diffusion LLM is more robust for the OOD position embeddings in length extrapolation. Therefore, unlike traditional auto-regressive LLMs that exhibit catastrophic performance degradation when exceeding their maximum supported context length, diffusion LLMs \textbf{\textit{maintain stable outputs}} and \textbf{\textit{demonstrate local perception}} in extended context.

\section{Context Extension For Diffusion LLMs}\label{obs_ntk}

\begin{figure}[!tb]
\begin{minipage}{0.98\textwidth}
    \begin{subfigure}[b]{0.48\linewidth}
        \centering
        \includegraphics[width=\linewidth]{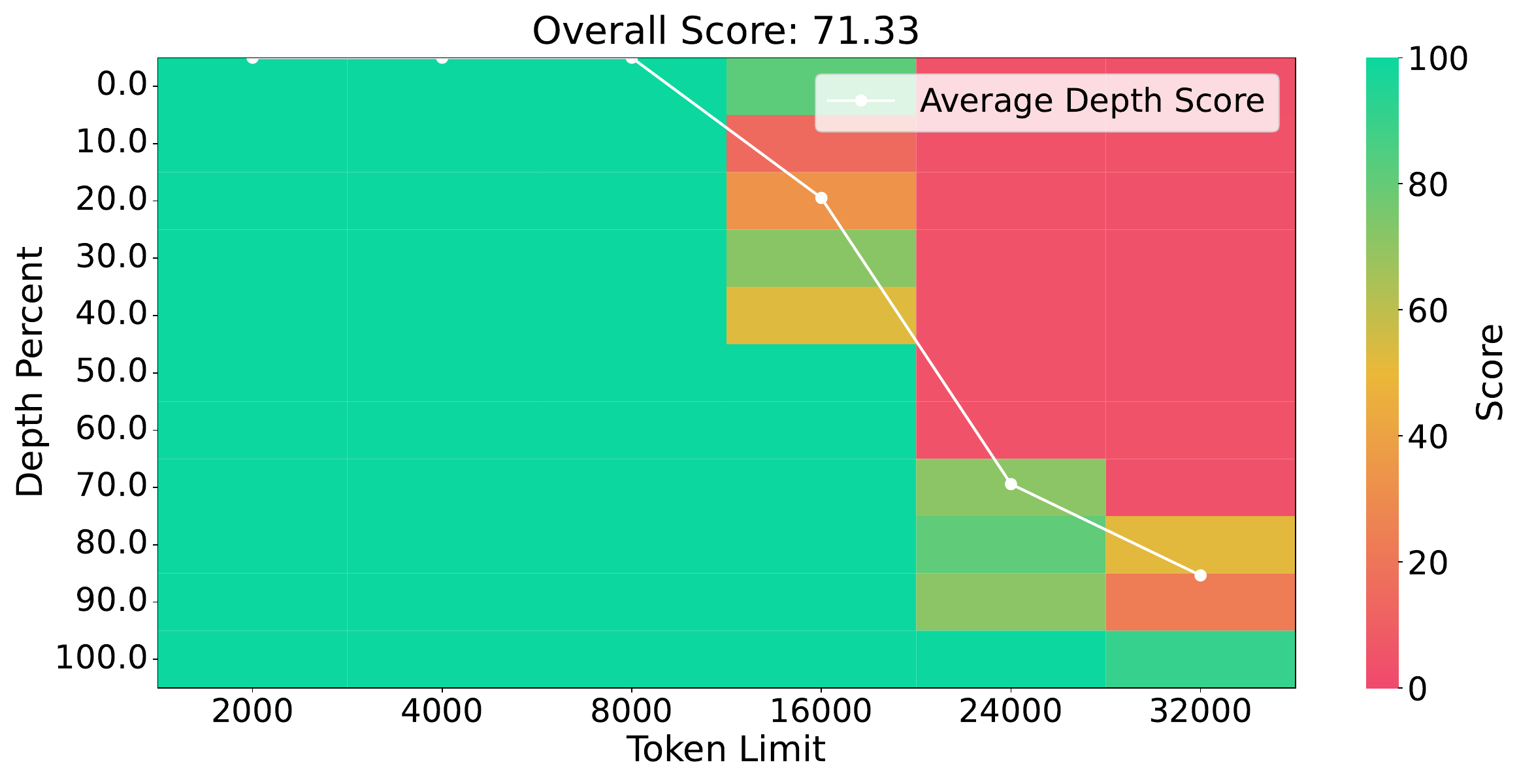}
        \caption{LLaDA-8B-Base with $\lambda=4$}
        \label{llada_8b_base_lambda4}
    \end{subfigure}
    \hfill
    \begin{subfigure}[b]{0.48\linewidth}
        \centering
        \includegraphics[width=\linewidth]{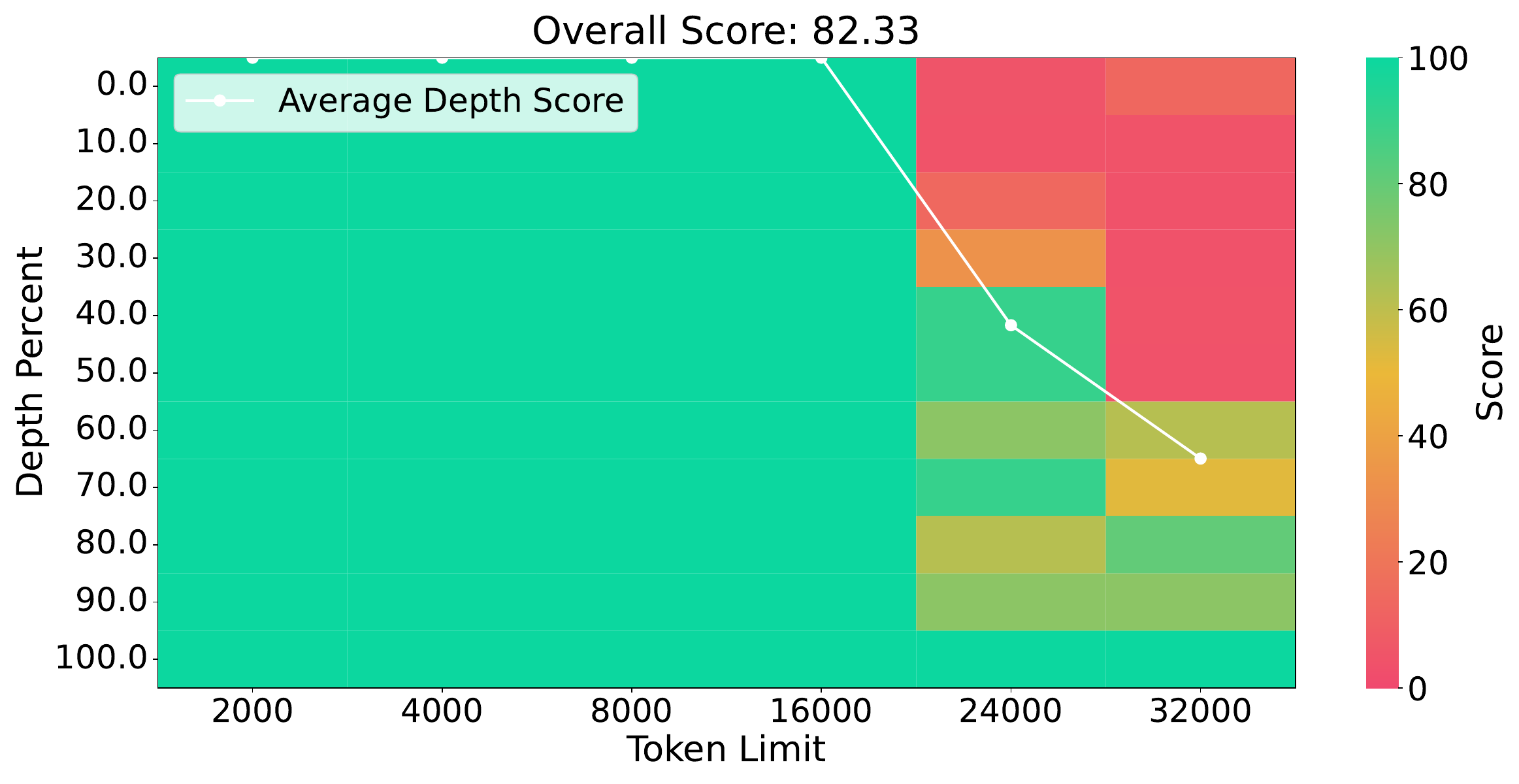}
        \caption{LLaDA-8B-Base with $\lambda=14$}
        \label{llada_8b_base_lambda14}
    \end{subfigure}
    \vskip\baselineskip
    \begin{subfigure}[b]{0.48\linewidth}
        \centering
        \includegraphics[width=\linewidth]{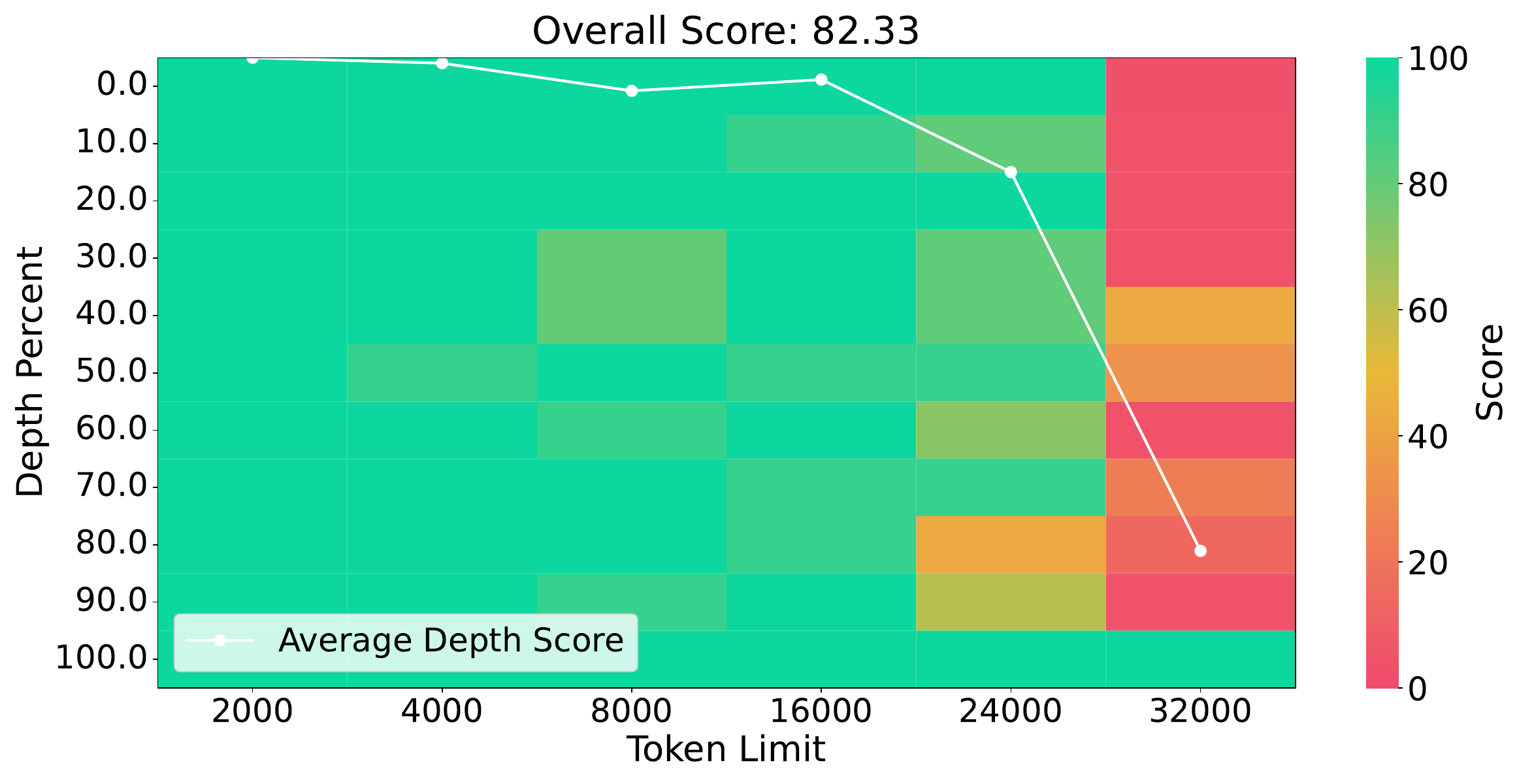}
        \caption{LLaDA-8B-Base with $\lambda=31$}
        \label{llada_8b_base_lambda31}
    \end{subfigure}
    \hfill
    \begin{subfigure}[b]{0.48\linewidth}
        \centering
        \includegraphics[width=\linewidth]{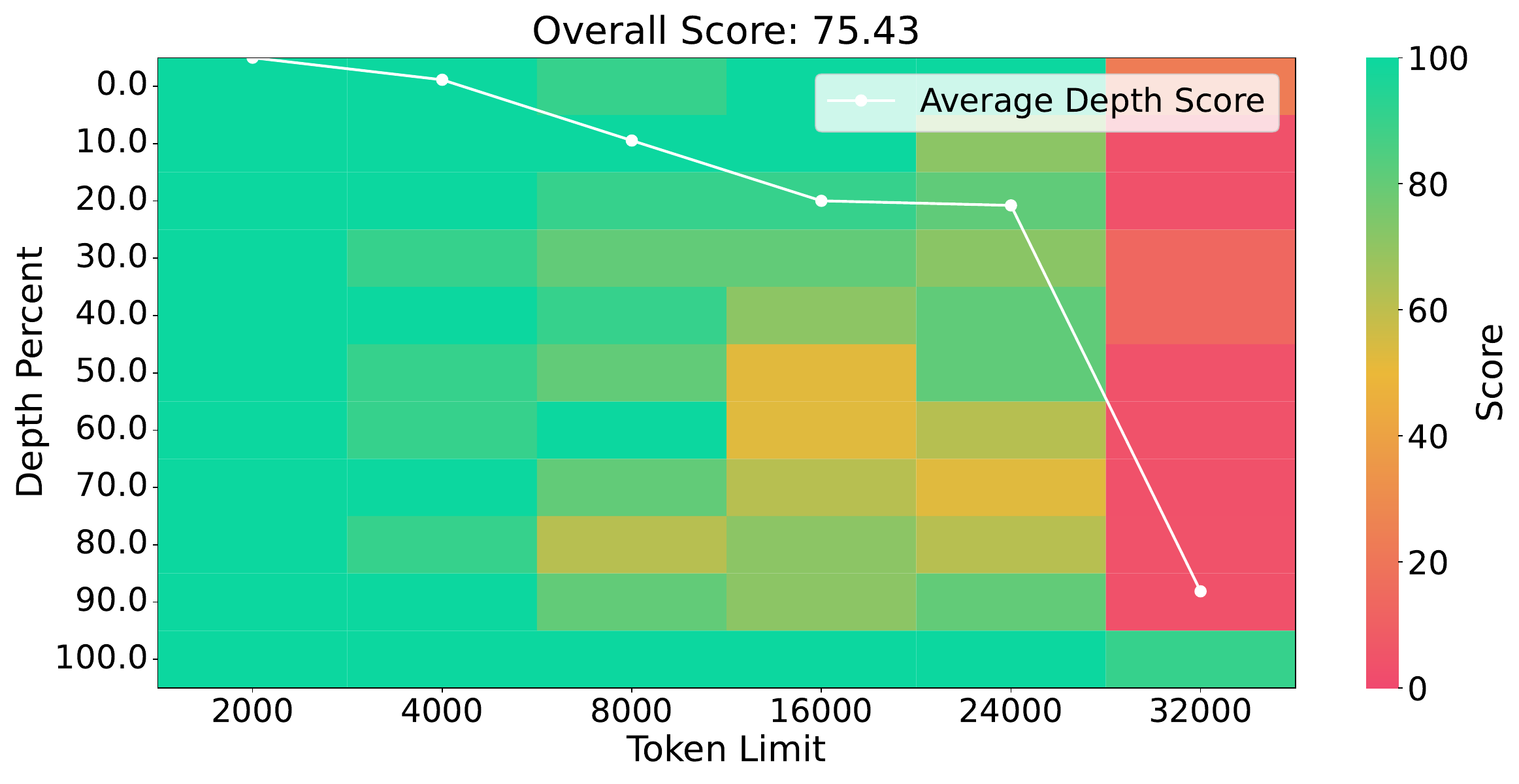}
        \caption{LLaDA-8B-Base with $\lambda=55$}
        \label{llada_8b_base_lambda55}
    \end{subfigure}
    \caption{NIAH Results of LLaDA-8B-Base~\citep{nie2025large} with different RoPE scaling factor.\label{llada_base_ntk}}
\end{minipage}
\end{figure}

\begin{figure}[!tb]
\begin{minipage}{0.98\textwidth}
    \begin{subfigure}[b]{0.48\linewidth}
        \centering
        \includegraphics[width=\linewidth]{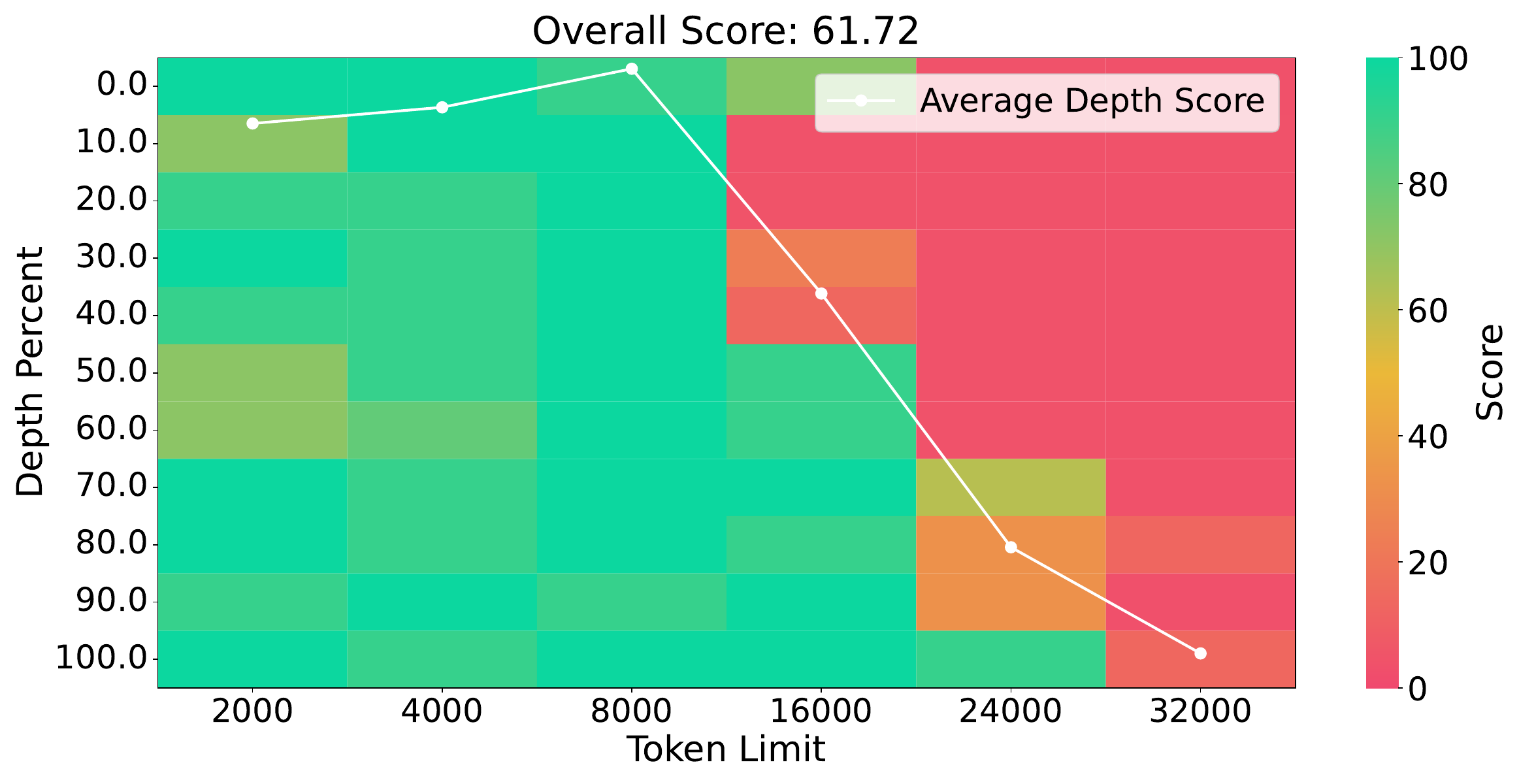}
        \caption{LLaDA-8B-Instruct with $\lambda=4$}
        \label{llada_8b_chat_lambda4}
    \end{subfigure}
    \hfill
    \begin{subfigure}[b]{0.48\linewidth}
        \centering
        \includegraphics[width=\linewidth]{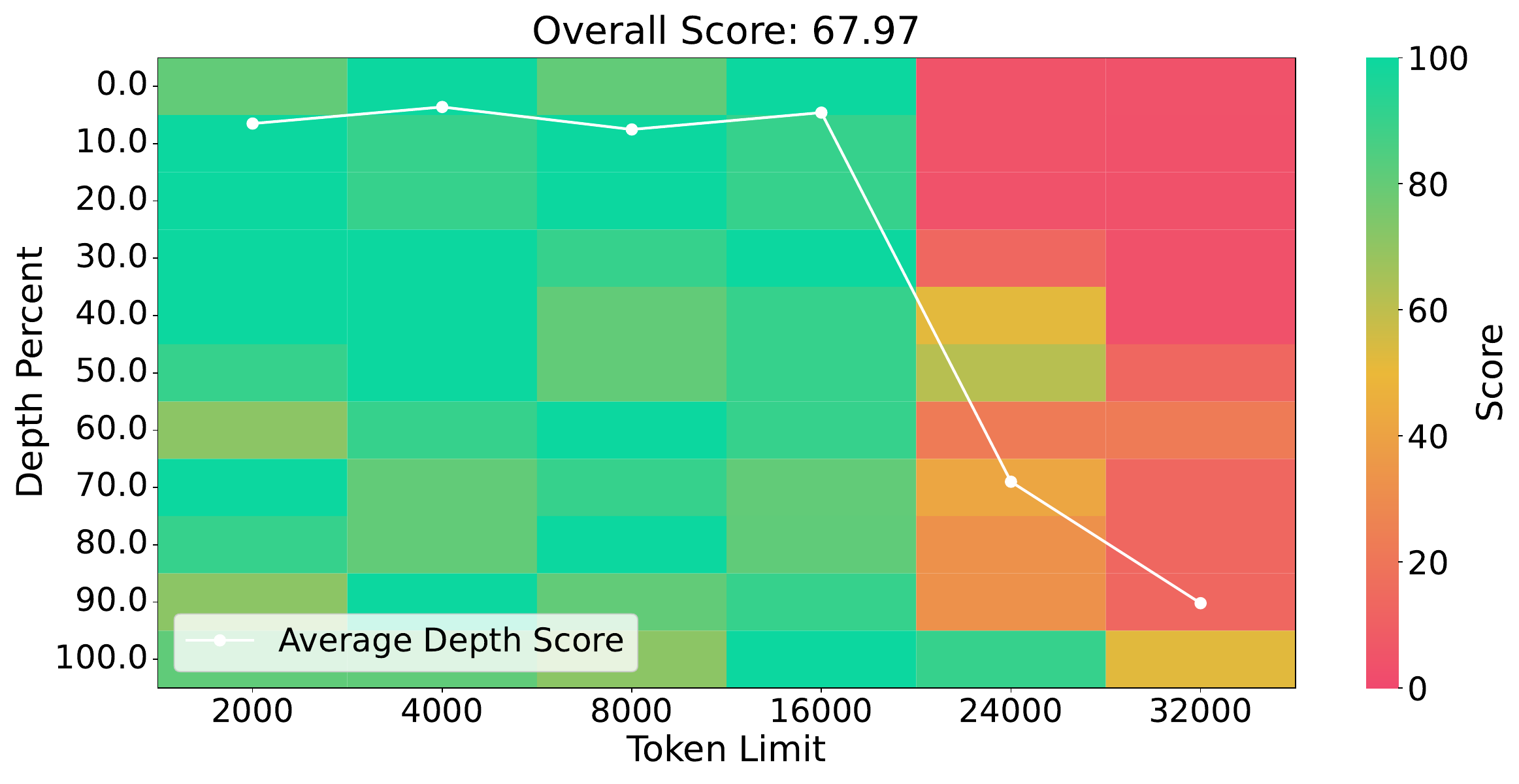}
        \caption{LLaDA-8B-Instruct with $\lambda=14$}
        \label{llada_8b_chat_lambda14}
    \end{subfigure}
    \vskip\baselineskip
    \begin{subfigure}[b]{0.48\linewidth}
        \centering
        \includegraphics[width=\linewidth]{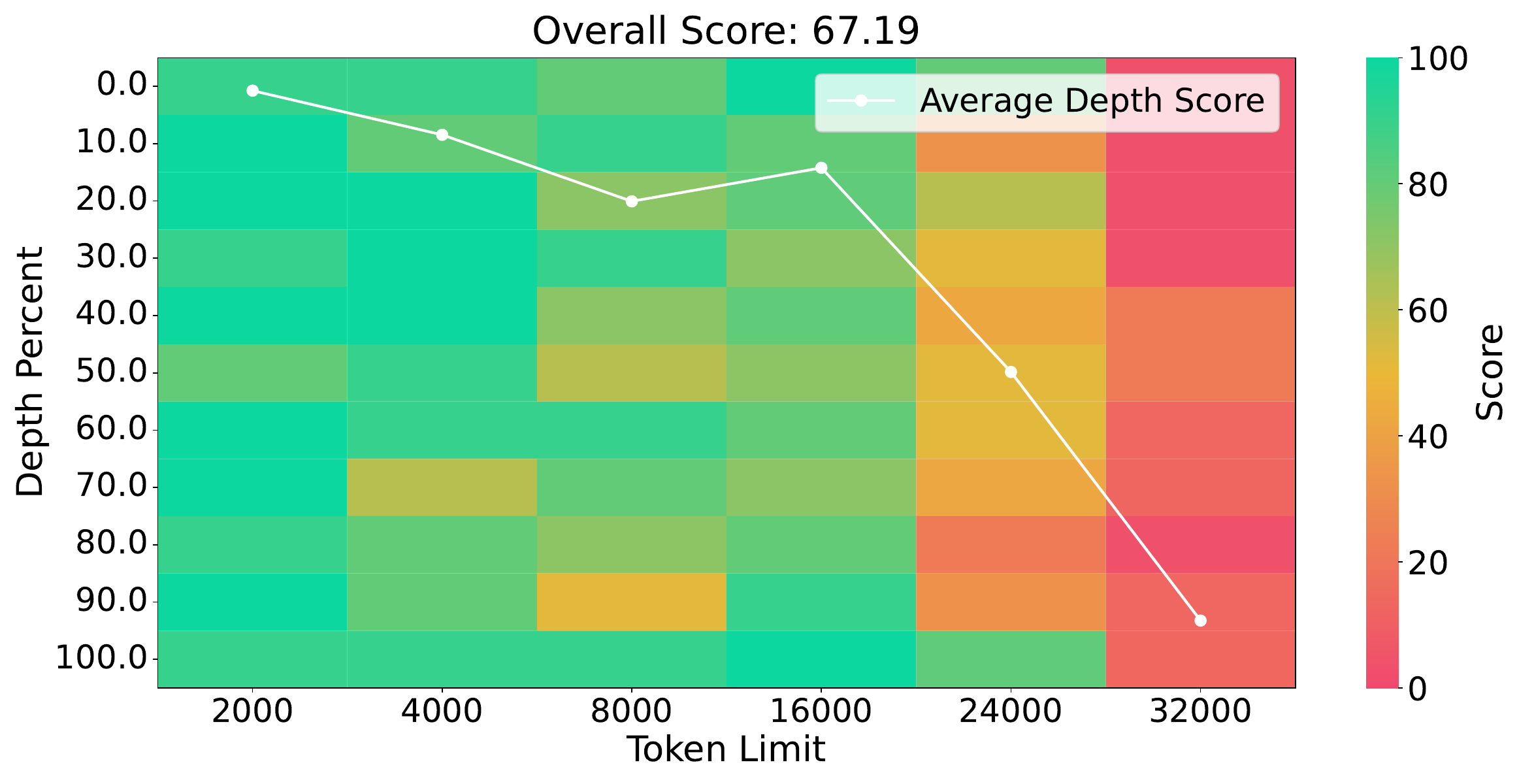}
        \caption{LLaDA-8B-Instruct with $\lambda=31$}
        \label{llada_8b_chat_lambda31}
    \end{subfigure}
    \hfill
    \begin{subfigure}[b]{0.48\linewidth}
        \centering
        \includegraphics[width=\linewidth]{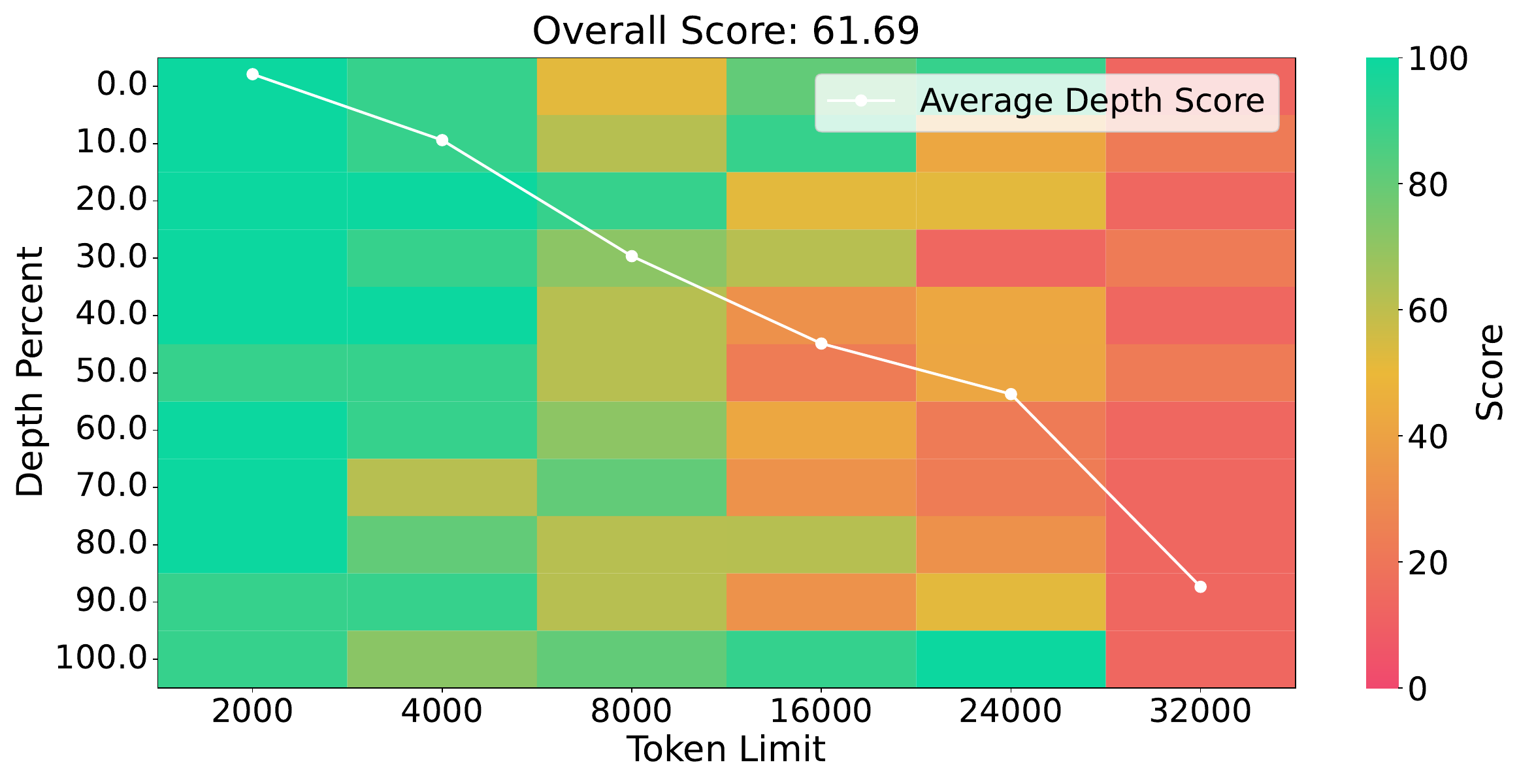}
        \caption{LLaDA-8B-Base with $\lambda=55$}
        \label{llada_8b_chat_lambda55}
    \end{subfigure}
    \caption{NIAH Results of LLaDA-8B-Instruct~\citep{nie2025large} with different RoPE scaling factor.\label{llada_chat_ntk}}
\end{minipage}
\end{figure}

Since the reason for the surprising phenomenon has been clarified, we now move on to the extrapolation methods for diffusion LLMs. Since the retrievable depth of diffusion LLMs remains constrained by the range of cosine values encountered during pre-training, we transfer the NTK-based extrapolation~\citep{fixedNTK} and its scaling laws~\citep{liu2023scaling} to diffusion LLMs, thus proposing the length extrapolation method for diffusion LLMs, LongLLaDA. As detailed in Appendix~\ref{preliminary}. The scaling factor $\lambda$ in training-free NTK scaling~\citep{fixedNTK} for RoPE-based auto-regressive LLMs is decided by the extrapolation context length $t$ and critical dimension $d_\text{extra}$ calculated by rotary base $\beta_0$ and pretrained context length $T_\text{train}$, as shown in Equation~\ref{equ_sum}.
\begin{equation}
\lambda={10}^{-4}\cdot{\left(\frac{t}{2\pi}\right)}^{d/d_\text{extra}},\quad d_\text{extra}=2\left\lceil\frac{d}{2}\log_{\beta_0}{\frac{T_\text{train}}{2\pi}}\right\rceil\text{.}\label{equ_sum}
\end{equation}
Similarly, in LongLLaDA, based on \citet{nie2025large}, the pretrained rotary base $\beta_0=500000$, and the pre-training context length $T_\text{train}$ is 4k. This yields a critical dimension $d_\text{extra}=64$. Accordingly, the required scaling factor $\lambda$ for extrapolation to 8k, 16k, 24k, and 32k is calculated as 4, 14, 31, and 55, respectively. The extrapolation results are illustrated in Figure~\ref{llada_base_ntk} and Figure~\ref{llada_chat_ntk}.

When $\lambda=4,14$, LongLLaDA can effectively extrapolate diffusion LLMs to the corresponding context lengths, achieving near 100\% recall across all depths within these ranges. As the context length increases beyond the extrapolation limit, the retrievable depth proportionally expands while maintaining the local-perception effect. The average depth score curves exhibit a right shift across different context lengths. When $\lambda=31$, a lost-in-the-middle phenomenon~\citep{liu2023lost} similar to auto-regressive models emerges in intermediate depths, indicating that LongLLaDA approaches its practical extrapolation limit~\citep{fixedNTK}. When $\lambda=55$, further extrapolation is unachievable. We also validate the effectiveness of LongLLaDA in LLaDA-1.5~\citep{zhu2025llada} in Appendix~\ref{more_results}. Consequently, for RoPE-based diffusion LLMs, \textbf{\textit{NTK extrapolation and its scaling law remain applicable}} during inference.

\section{Task-Driven Long-Context Capability Analysis}\label{obs_task}

\begin{table}[!tb]
\tabcolsep=0.1cm
\centering\small
\begin{tabular}{lcccccccccccccc}
\toprule
& \multicolumn{7}{c}{\textbf{4k}} & \multicolumn{7}{c}{\textbf{8k}} \\
\cmidrule(lr){2-8} \cmidrule(lr){9-15} 
& SD & MD & Sum & ICL & Syn & Code & Avg & SD & MD & Sum & ICL & Syn & Code & Avg \\
\midrule
\textbf{\textit{LLaDA-8B-Base}} & 15.1 & 18.4 & 32.0 & \textbf{42.0} & 54.7 & 59.6 & 34.1 & \cellcolor{gray!25}13.9 & \cellcolor{gray!25}13.1 & \cellcolor{gray!25}30.8 & \cellcolor{gray!25}40.7 & \cellcolor{gray!25}56.0 & \cellcolor{gray!25}57.4 & \cellcolor{gray!25}32.4 \\
+  $\lambda=4$ & 14.7 & 19.1 & 31.5 & 40.9 & 52.4 & 63.0 & 34.0 & 15.2 & 18.6 & 31.0 & \underline{41.4} & 53.8 & 59.2 & 33.6 \\
\midrule
\textbf{\textit{LLaDA-8B-Instruct}} & \underline{25.1} & 19.4 & 30.6 & 36.4 & 62.8 & 62.7 & 37.2 & \cellcolor{gray!25}22.7 & \cellcolor{gray!25}14.0 & \cellcolor{gray!25}33.4 & \cellcolor{gray!25}33.2 & \cellcolor{gray!25}66.7 & \cellcolor{gray!25}66.4 & \cellcolor{gray!25}36.8 \\
+  $\lambda=4$ & 22.1 & \underline{19.8} & 33.0 & 38.0 & 63.3 & 65.0 & \textbf{37.8} & \underline{23.4} & 19.8 & \textbf{35.3} & 39.8 & \textbf{72.9} & 67.3 & 40.6 \\
\midrule
\textbf{\textit{LLaDA-1.5}} & 24.4 & 19.4 & 31.6 & 33.5 & \textbf{63.6} & \underline{66.7} & 37.6 & \cellcolor{gray!25}22.6 & \cellcolor{gray!25}14.5 & \cellcolor{gray!25}33.4 & \cellcolor{gray!25}33.0 & \cellcolor{gray!25}67.6 & \cellcolor{gray!25}67.6 & \cellcolor{gray!25}37.1 \\
+  $\lambda=4$ & 21.8 & 19.7 & \underline{33.1} & 35.3 & \underline{63.4} & \textbf{67.3} & \textbf{37.8} & 23.0 & \underline{20.6} & \underline{34.9} & 39.3 & \textbf{72.9} & \underline{67.9} & \underline{40.7} \\
\midrule
\textbf{\textit{LLaMA3-8B-Base}} & 17.2 & 18.7 & 25.0 & \underline{41.7} & 47.6 & 66.5 & 33.6 & 18.2 & 18.3 & 26.1 & \textbf{44.5} & 49.6 & \textbf{69.4} & 35.1 \\
\midrule
\textbf{\textit{LLaMA3-8B-Instruct}} & \textbf{31.9} & \textbf{26.1} & \textbf{33.6} & 39.6 & 46.6 & 55.9 & 37.0 & \textbf{37.5} & \textbf{28.3} & 34.7 & 40.7 & 62.8 & 56.1 & \textbf{41.9} \\
\bottomrule
\end{tabular}
\caption{Results of LLaDA-8B~\citep{nie2025large}, LLaDA-1.5~\citep{zhu2025llada} and LLaMA3-8B~\citep{meta2024llama} on LongBench~\citep{bai2023longbench} under 4k and 8k context length. Gray cells indicate that the evaluation context length exceeds the context length supported by the evaluated LLM. SD, MD, Sum, and Syn stand for Single-Doc QA, Multi-Doc QA, Summarization, and Synthetic tasks, while Avg is the average score of all subtasks weighted by the evaluation data number.\label{tab_longbench}}
\end{table}

\begin{table}[!tb]
\tabcolsep=0.12cm
\centering\small
\begin{tabular}{lcccccccccccc}
\toprule
& \multicolumn{4}{c}{\textbf{4k}} & \multicolumn{4}{c}{\textbf{8k}} & \multicolumn{4}{c}{\textbf{16k}} \\
\cmidrule(lr){2-5} \cmidrule(lr){6-9} \cmidrule(lr){10-13} 
~ & NIAH & AGG & QA & Avg & NIAH & AGG & QA & Avg & NIAH & AGG & QA & Avg \\ 
\midrule
\textbf{\textit{LLaDA-8B-Base}} & 99.7 & 65.2 & 82.5 & 89.1 & \cellcolor{gray!25}53.8 & \cellcolor{gray!25}45.2 & \cellcolor{gray!25}41.0 & \cellcolor{gray!25}49.8 & \cellcolor{gray!25}22.0 & \cellcolor{gray!25}1.9 & \cellcolor{gray!25}36.0 & \cellcolor{gray!25}19.5 \\ 
+  $\lambda=4$ & 99.5 & 82.3 & 80.5 & 92.6 & 96.4 & 61.0 & 73.0 & 84.7 & \cellcolor{gray!25}51.8 & \cellcolor{gray!25}17.1 & \cellcolor{gray!25}53.5 & \cellcolor{gray!25}44.1 \\ 
+  $\lambda=14$ & 99.8 & 83.5 & 77.0 & 92.5 & 99.3 & 68.9 & 64.0 & 86.8 & 85.4 & 48.1 & 54.0 & 72.0 \\ 
+  $\lambda=31$ & \textbf{100.0} & 83.8 & 77.0 & 92.7 & 97.8 & 75.2 & 62.5 & 87.1 & \textbf{97.2} & 51.8 & 41.0 & 78.0 \\ 
\midrule
\textbf{\textit{LLaDA-8B-Instruct}} & 99.3 & 57.8 & \underline{90.5} & 88.4 & \cellcolor{gray!25}52.3 & \cellcolor{gray!25}44.3 & \cellcolor{gray!25}48.0 & \cellcolor{gray!25}49.8 & \cellcolor{gray!25}18.9 & \cellcolor{gray!25}12.0 & \cellcolor{gray!25}47.0 & \cellcolor{gray!25}21.6 \\ 
+  $\lambda=4$ & 99.8 & 65.5 & 89.5 & 90.3 & 95.9 & 56.6 & 89.0 & 85.8 & \cellcolor{gray!25}41.6 & \cellcolor{gray!25}31.0 & \cellcolor{gray!25}73.0 & \cellcolor{gray!25}44.0 \\ 
+  $\lambda=14$ & \textbf{100.0} & 76.4 & 89.0 & 92.9 & 97.3 & 66.2 & \textbf{89.5} & 88.9 & 67.1 & 53.8 & \textbf{84.5} & 66.7 \\ 
+  $\lambda=31$ & \textbf{100.0} & 77.7 & 86.5 & 92.8 & 98.8 & 73.5 & 88.5 & 91.3 & 88.0 & 62.2 & 82.0 & 81.1 \\ 
\midrule  
\textbf{\textit{LLaDA-1.5}} & 98.7 & 66.0 & 90.0 & 89.8 & \cellcolor{gray!25}53.9 & \cellcolor{gray!25}45.1 & \cellcolor{gray!25}48.5 & \cellcolor{gray!25}51.0 & \cellcolor{gray!25}19.0 & \cellcolor{gray!25}14.2 & \cellcolor{gray!25}46.0 & \cellcolor{gray!25}22.1 \\
+  $\lambda=4$ & 99.8 & 73.9 & \textbf{91.0} & 92.5 & 96.3 & 59.3 & 88.0 & 86.5 & \cellcolor{gray!25}43.3 & \cellcolor{gray!25}32.2 & \cellcolor{gray!25}73.0 & \cellcolor{gray!25}45.3 \\
+  $\lambda=14$ & \textbf{100.0} & 79.8 & 88.5 & 93.6 & \textbf{99.9} & 67.8 & \underline{89.0} & 90.8 & 67.4 & 51.6 & \underline{84.0} & 66.3 \\
+  $\lambda=31$ & \textbf{100.0} & 81.6 & 87.5 & 93.8 & 98.9 & 75.1 & 86.5 & 91.5 & 85.8 & 58.2 & 81.5 & 78.7 \\
\midrule
\textbf{\textit{LLaMA3-8B-Base}} & 99.8 & 98.1 & 67.5 & \underline{94.4} & 99.6 & 93.5 & 63.0 & 92.5 & \cellcolor{gray!25}0.0 & \cellcolor{gray!25}0.0 & \cellcolor{gray!25}0.0 & \cellcolor{gray!25}0.0 \\
+  $\lambda=4$ & 99.9 & \textbf{98.7} & 65.0 & 94.2 & \underline{99.8} & \textbf{94.1} & 59.0 & \underline{92.2} & \underline{97.0} & 86.6 & 54.5 & \underline{88.1} \\
+  $\lambda=13$ & 99.5 & \underline{98.6} & 66.0 & 94.1 & 99.1 & \underline{94.0} & 59.0 & 91.8 & 93.8 & \textbf{90.3} & 56.0 & 87.2 \\
\midrule
\textbf{\textit{LLaMA3-8B-Instruct}} & 99.6 & 97.2 & 68.5 & 94.3 & 98.2 & 92.6 & 54.0 & 90.1 & \cellcolor{gray!25}0.0 & \cellcolor{gray!25}0.0 & \cellcolor{gray!25}0.0 & \cellcolor{gray!25}0.0 \\
+  $\lambda=4$ & 99.8 & 96.9 & 72.0 & \textbf{94.9} & 99.6 & 93.5 & 65.0 & \textbf{92.8} & 95.0 & \underline{89.5} & 63.0 & \textbf{88.8} \\
+  $\lambda=13$ & 99.5 & 96.7 & 68.0 & 94.0 & 99.3 & 92.4 & 63.5 & \underline{92.2} & 95.3 & 78.6 & 62.5 & 86.4 \\
\bottomrule
\end{tabular}
\caption{Results of LLaDA-8B~\citep{nie2025large}, LLaDA-1.5~\citep{zhu2025llada} and LLaMA3-8B~\citep{meta2024llama} on RULER~\citep{hsieh2024ruler} under 4k, 8k and 16k context length. \label{tab_ruler}}
\end{table}

Regarding the downstream long-context performance of diffusion LLMs and their difference from traditional auto-regressive LLMs, apart from the NIAH retrieval evaluation, we conduct comparative analyses across more benchmarks using LLaDA and LLaMA as examples. We first evaluate LLaDA-8B~\citep{nie2025large}, LLaDA-1.5~\citep{zhu2025llada}, and LLaMA3-8B~\citep{meta2024introducing}, including pre-trained models and those employing NTK-based extrapolation during inference, with LongBench~\citep{bai2023longbench}, in 4k and 8k context length, with the exceeding part being truncated from the middle. For the summary tasks, the output length is 512, while for the others, the output length is 64. We still keep the sampling steps the same as the output length, and the block size to 64 for diffusion LLMs. The results are shown in Table~\ref{tab_longbench}. Still, LLaDA can give a stable output and get a decent performance beyond the maximum supported context length. Moreover, we find that in all task domains besides synthetic tasks, the difference between LLaDA Series and LLaMA3 Series is relatively limited compared with the difference within LLaMA3 Series. Only in the synthetic domain do LLaDA Series outperform LLaMA3 Series consistently. This inspires us to conduct an in-depth discussion of diffusion LLMs on the performance of the synthesis tasks compared with auto-regressive LLMs. 

We further the discussion with RULER benchmark~\citep{hsieh2024ruler}, we compare LLaDA-8B~\citep{nie2025large}, LLaDA-1.5~\citep{zhu2025llada}, and LLaMA3-8B~\citep{meta2024introducing}, at context lengths of 4k, 8k, and 16k. We set the block size and sampling steps to 64 for diffusion LLMs. The results are shown in Table~\ref{tab_ruler}. First, consistent with the NIAH results, auto-regressive LLMs fail to produce valid outputs beyond their effective context length, while diffusion LLMs maintain measurable performance. Regarding task types, diffusion LLMs achieve comparable results to auto-regressive LLMs on NIAH tasks, including Single-Key, Multi-Key, Multi-Query, and Multi-Value variants. However, diffusion LLMs show significantly inferior performance in aggregation tasks, including Variable Tracing and Frequent or Common Word Extraction, where auto-regressive LLMs typically perform well. Surprisingly, on QA tasks, including SQuAD and Hotpot, that challenge auto-regressive LLMs~\citep{hsieh2024ruler}, diffusion LLMs demonstrate superior capability. These observations reveal the distinctive characteristics of diffusion LLMs in long-context tasks, that current diffusion LLMs, like LLaDA, demonstrate comparable performance to the auto-regressive LLMs, like LLaMA3, in most task types, but \textbf{\textit{underperform in aggregation tasks}}, and \textbf{\textit{outperform in synthetic QA tasks}} consistently.

\section{Related Work}\label{related}

\paragraph{Large Language Diffusion Models} Recently, Large Language Diffusion Models, or diffusion LLMs, have become a widely discussed topic in NLP research. After the theoretical simplification~\citep{sahoo2024simple,ou2024your,shi2024simplified} and empirical verification~\citep{he2022diffusionbert,gong2024scaling}, researchers scale the size of diffusion LLMs to billions of parameters~\citep{nie2024scaling,nie2025large,dream2025} and demonstrate that diffusion LLMs can achieve comparable results with more promising performance in the reversal curse~\citep{berglund2023reversal}. These immediately attract the attention of many more researchers. Significant research efforts have focused on adapting diffusion LLMs for multimodality, such as MMaDA~\citep{yang2025mmada}, LLaDA-V~\citep{you2025llada}, and LaViDa~\citep{li2025lavida}, applying them to reasoning tasks, such as d1~\citep{zhao2025d1}, DCoLT\citet{huang2025reinforcing}, and LLaDA-1.5~\citep{zhu2025llada}, and optimizing their efficiency~\citep{ma2025dkv,hu2025accelerating,wu2025fast}, including dKV-Cache~\citep{ma2025dkv}, Dimple~\citep{yu2025dimple}, dLLM-Cache~\citep{liudllm}, FreeCache~\citep{hu2025accelerating}, Fast-dLLM~\citep{wu2025fast}, and so on. However, there is still no discussion on the long-context capability of diffusion LLMs.

\paragraph{Length Extrapolation in LLM} Length extrapolation, or length generalization, or context extension, is an important issue for LLMs~\citep{press2021train,liu2025thus}. The mainstream extrapolation research mainly focuses on adjusting position embedding, especially the widely used RoPE~\citep{su2021roformer}. For example, Linear PI~\citep{chen2023extending} first achieves LLMs' length extrapolation by scaling position indices to the pre-training range with little fine-tuning. The NTK method~\citep{fixedNTK,dynamicNTK,peng2023yarn} then scales the rotary base in RoPE~\citep{su2021roformer} to achieve plug-and-play length extrapolation. Subsequently, amplifying the rotary base and training on longer lengths has become the dominant approach for length extrapolation~\citep{roziere2023code,xiong2023effective,liu2023scaling,ding2024longrope}. In addition, ReRoPE~\citep{rerope}, ReAttention~\citep{liu2024reattention}, and DCA~\citep{an2024training,an2024does} also achieve plug-and-play extrapolation by limiting the relative position. In this paper, we still focus on the length extrapolation via NTK scaling~\citep{fixedNTK,liu2023scaling} in the inference stage, and try to reveal and explain the similarities and differences in length extrapolation between diffusion-based and auto-regressive LLM.

\section{Conclusion}

In this work, we provide the first systematic analysis of long-context capabilities in diffusion LLMs. We demonstrate and analyze their characteristics for stable perplexity and local perception in direct context extrapolation from the perspective of the RoPE dynamic. Then, we propose LongLLaDA, which extends the context length in NTK scaling effectively without further training, and validate that the scaling laws still work for diffusion LLMs. Besides, we also show that diffusion LLMs match auto-regressive models on the average score of LongBench as well as the retrieval tasks, lag in aggregation tasks, but excel at QA in RULER evaluation. We hope our work can pave the foundation for future long-context research in diffusion LLMs.

\section*{Limitation}

Although we have conducted extensive experiments on diffusion LLMs, our results mainly focus on LLaDA Series and the inference stage. We will carry out the fine-tuning extrapolation experiments in the future to verify more conclusions in RoPE-based scaling theory for auto-regressive LLMs. Besides, we will also add more analyses focused on sampling strategy specialized in diffusion LLMs.




\bibliography{iclr2025_conference}
\bibliographystyle{iclr2025_conference}

\appendix

\section{Preliminary: RoPE Extrapolation in Auto-regressive LLM}\label{preliminary}

Rotary Position Embedding (RoPE)~\citep{su2021roformer} employs trigonometric functions to encode absolute positions in Q state $\bm{q}_t=\left[q_t^{(0)},\cdots,q_t^{(d-1)}\right]$ and K state $\bm{k}_s=\left[k_s^{(0)},\cdots,k_s^{(d-1)}\right]$. By leveraging the properties of rotation matrices, RoPE encodes relative position in the attention matrix $\bm{A}$, as shown in Equation~\ref{equ_rope_def} and demonstrates superior performance, thus being widely adopted by many auto-regressive LLMs~\citep{Sun2024MOSS,dubey2024llama,qwen2}.
\begin{equation}
\begin{aligned}
\bm{A}_{t,s}&=\left(\bm{q}_t\bm{R}_t\right)\left(\bm{k}_s\bm{R}_s\right)^\top=\bm{q}_t\bm{R}_{t-s}\bm{k}_s^\top \\
&=\sum\limits_{n=0}^{d/2-1}{\begin{bmatrix}q_t^{(2n)}\\q_t^{(2n+1)}\end{bmatrix}^\top \begin{bmatrix}\cos{\theta_n(t-s)}&-\sin{\theta_n(t-s)}\\\sin{\theta_n(t-s)}&\cos{\theta_n(t-s)}\end{bmatrix} \begin{bmatrix}k_s^{(2n)}\\k_s^{(2n+1)}\end{bmatrix}} \\
&=\sum\limits_{n=0}^{d/2-1}{\begin{gathered}\left(q_t^{(2n)}k_s^{(2n)}+q_t^{(2n+1)}k_s^{(2n+1)}\right)\cos{\theta_n(t-s)}\\-\left(q_t^{(2n)}k_s^{(2n+1)}-q_t^{(2n+1)}k_s^{(2n)}\right)\sin{\theta_n(t-s)}
\end{gathered}}
\end{aligned}\text{.}\label{equ_rope_def}\end{equation}
However, RoPE still faces the length extrapolation issue~\citep{press2021train}. When RoPE-based auto-regressive LLMs are tested beyond the pre-trained context length, the perplexity rises significantly, and downstream performance drops sharply. The underlying causes and corresponding solutions can be attributed to two key properties of trigonometric functions: \textbf{\textit{periodicity}} and \textbf{\textit{monotonicity}}.

\paragraph{Rule of Periodicity} According to the design of RoPE, different dimensions of $\bm{q}_t,\bm{k}_s$ use different rotary angles $\theta_n$, with rotary base $\beta_0=10000$ by default and the periods $T_n$ for $\sin(\theta_nt)$ and $\cos(\theta_nt)$ increasing from low to high dimensions as shown in Equation~\ref{equ_theta}.
\begin{equation}
    \theta_n=\beta_0^{-2n/d}, \quad T_n=2\pi\cdot\beta_0^{2n/d}, \quad n=0,\cdots,d/2-1
\text{.}\label{equ_theta}\end{equation}
For lower dimensions, $T_n$ is very short, compared with the pre-trained context length $T_\text{train}$, while for higher ones, $T_n$ becomes significantly longer, exceeding $T_\text{train}$. Consequently, there exists a \textit{\textbf{critical dimension}}, $d_\text{extra}$, as shown in Equation~\ref{equ_d_extra}, within which $\sin(\theta_nt)$ or $\cos(\theta_nt)$ complete at least one full period within the pretrained length, whereas those beyond do not.
\begin{equation}
d_\text{extra}=2\left\lceil\frac{d}{2}\log_{\beta_0}{\frac{T_\text{train}}{2\pi}}\right\rceil
\text{.}\label{equ_d_extra}\end{equation}
Therefore, dimensions beyond $d_\text{extra}$ will encounter OOD position embedding when processing longer inputs and larger position indices in inference, leading to extrapolation issues~\citep{liu2023scaling}.

To enable LLM to handle unseen position indices, NTK methods~\citep{fixedNTK,xiong2023effective} scale the rotary base by a factor $\lambda$, reducing the rotary angle to achieve position interpolation. However, since different dimensions undergo different degrees of interpolation, the position embedding at the critical dimension will first become OOD. Thus, based on the scaled period of the critical dimension, the extrapolation upper bound $T_\text{extra}$ for NTK methods can be derived, as shown in Equation~\ref{equ_t_extra}.
\begin{equation}
T_\text{extra}=2\pi\cdot{\left(\lambda\cdot\beta_0\right)}^{d_\text{extra}/d}
\text{.}\label{equ_t_extra}\end{equation}

Based on Equation~\ref{equ_t_extra}, for an input length $t$, the rotary base scaling factor $\lambda$ should be set as shown in Equation~\ref {equ_scale} to ensure no OOD position embeddings occur. Notably, this adjustment coefficient exhibits a sup-linear, power-law increase with inference length~\citep{liu2023scaling,liu2025thus}.
\begin{equation}
\lambda_t=\beta_0^{-1}\cdot{\left(\frac{t}{2\pi}\right)}^{d/d_\text{extra}}
\text{.}\label{equ_scale}\end{equation}
It should be noted that while such interpolation could theoretically avoid extrapolation issues, it can only achieve 2$\times$ to 6$\times$ long-context extension during inference, as longer inputs lead to increased attention entropy, limiting further extrapolation~\citep{fixedNTK,han2023lm,wang2024length}.

\paragraph{Rule of Monotonicity}

Since the pre-trained position information of dimensions beyond the critical dimension limits the extrapolation capability of RoPE-based auto-regressive LLMs, if the rotary base is reduced, and each dimension can cover at least half or even a full period, the perplexity curve of auto-regressive LLMs will be flattened~\citep{liu2023scaling}. However, this does not imply real length extrapolation. Subsequent studies~\citep{men2024base,hu2024can} find that such LLMs can only perceive local information in downstream evaluations and fail to retrieve long-context dependencies.

Exposing LLM to periodic position information leads to downstream degeneration, manifesting a sliding-window effect, which reveals another aspect of RoPE-based extrapolation, the impact of monotonicity. Although higher dimensions do not observe complete position information, they provide a relatively complete monotonic interval, reflecting partial ordering in long-context scenarios. These dimensions exhibit larger activation values in long-context tasks~\citep{jin2025massive}, are more sensitive to modeling long-context dependencies~\citep{liu2024beyond}, and are better suited for capturing sequential relationships~\citep{wei2025videorope}. Thus, solely optimizing for periodicity at the cost of losing monotonicity across all dimensions is wrong~\citep{men2024base,liu2025thus}.

\section{More Experiment Results}\label{appendix}

\subsection{Setup}\label{setup}

Since we have clarified the RoPE-based extrapolation in auto-regressive LLM, we now turn our focus to that in diffusion LLM and try to answer the three questions raised in Section~\ref{intro}. We conduct experiments on the existing diffusion LLM series, including LLaDA-8B~\citep{nie2025large}, LLaDA-1.5~\citep{zhu2025llada}, and Dream-v0~\citep{dream2025}. By default, we set the number of sampling steps in diffusion LLM to 32 and keep the sampling strategy in the official code of LLaDA~\footnote{https://github.com/ML-GSAI/LLaDA} and Dream~\footnote{https://github.com/HKUNLP/Dream}. We use OpenCompass~\citep{2023opencompass} for validation. All experiments are performed with a fixed random seed of 2025, FP16 precision for LLaDA Series and BF16 for Dream Series, and accelerated with FlashAttention2~\citep{dao2023flashattention}.

\subsection{More Experiment Results}\label{more_results}

\begin{figure}[!tb]
\begin{minipage}{0.98\textwidth}
    \begin{subfigure}[b]{0.48\linewidth}
        \centering
        \includegraphics[width=\linewidth]{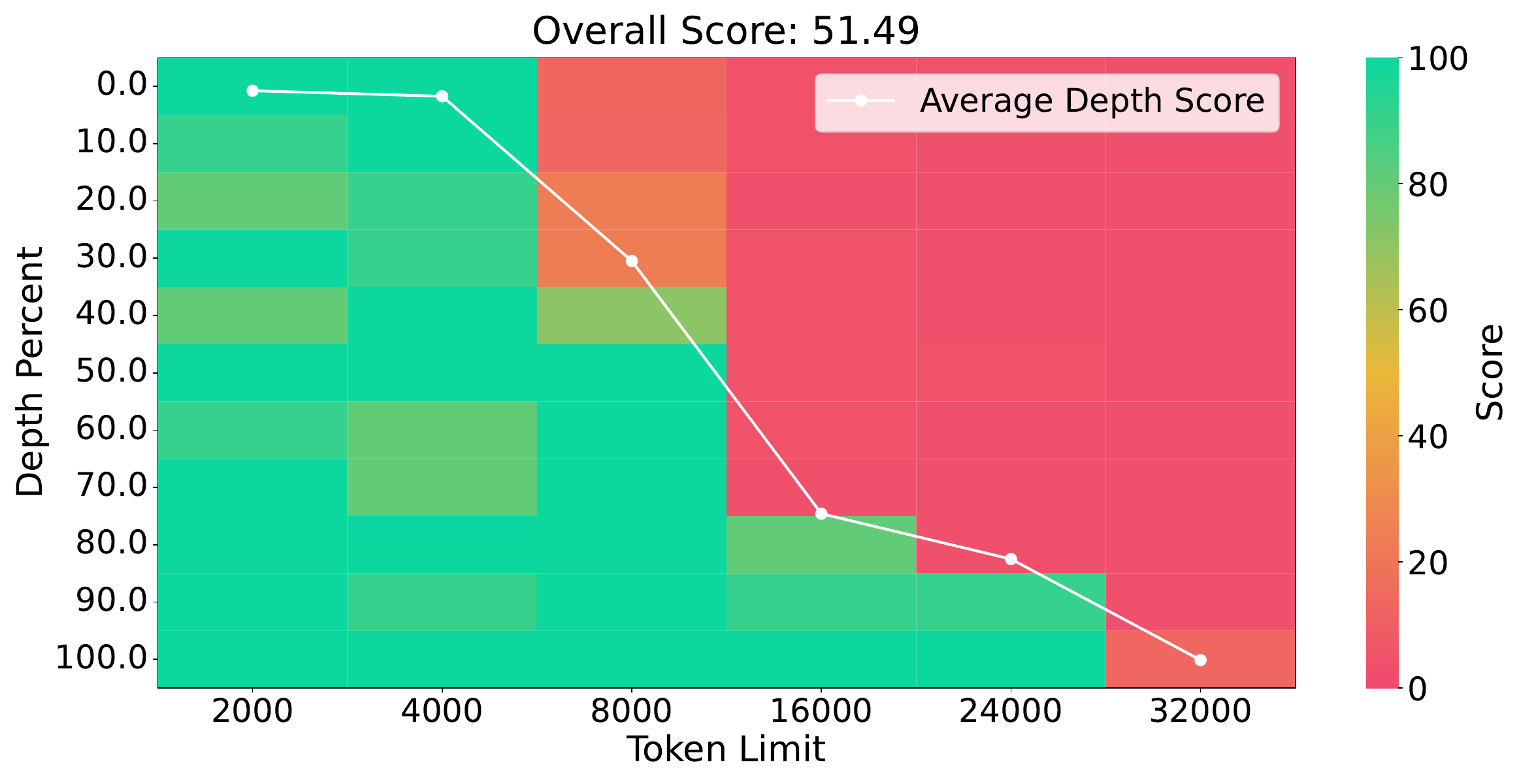}
        \caption{Pretrained LLaDA-1.5}
        \label{llada_1_5_niah}
    \end{subfigure}
    \hfill
    \begin{subfigure}[b]{0.48\linewidth}
        \centering
        \includegraphics[width=\linewidth]{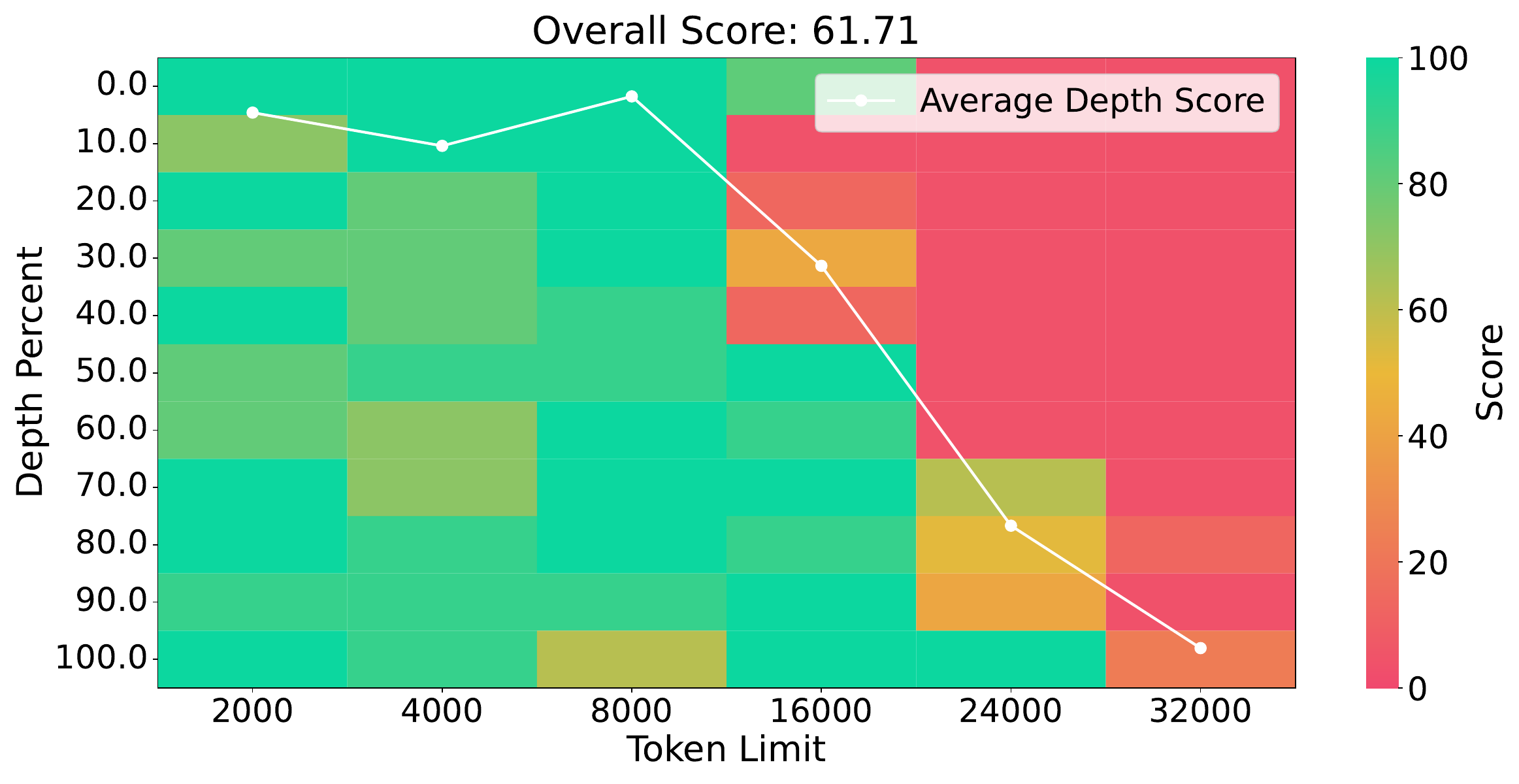}
        \caption{LLaDA-1.5 with $\lambda=4$}
        \label{llada_1_5_niah_lambda4}
    \end{subfigure}
    \vskip\baselineskip
    \begin{subfigure}[b]{0.48\linewidth}
        \centering
        \includegraphics[width=\linewidth]{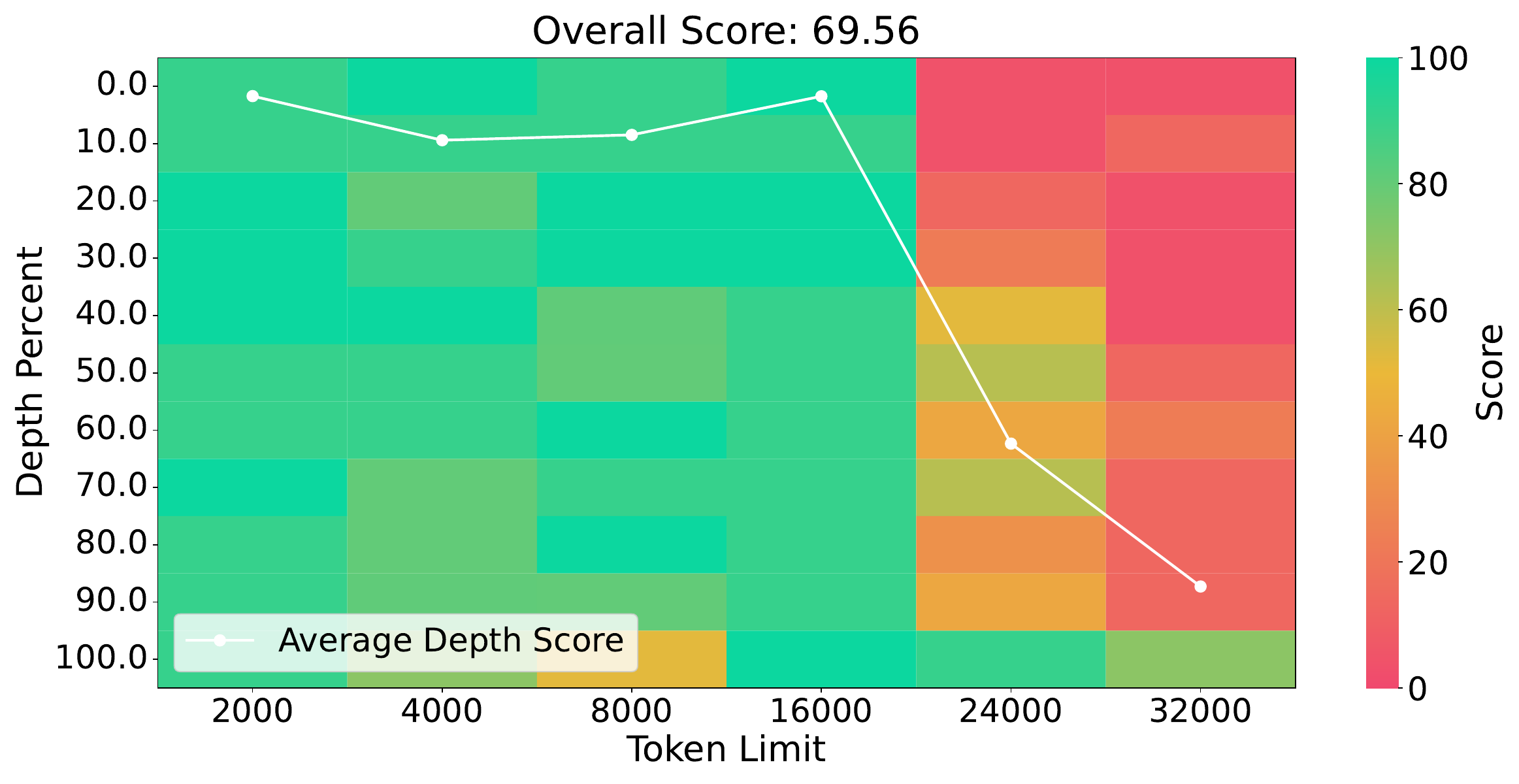}
        \caption{LLaDA-1.5 with $\lambda=14$}
        \label{llada_1_5_niah_lambda14}
    \end{subfigure}
    \hfill
    \begin{subfigure}[b]{0.48\linewidth}
        \centering
        \includegraphics[width=\linewidth]{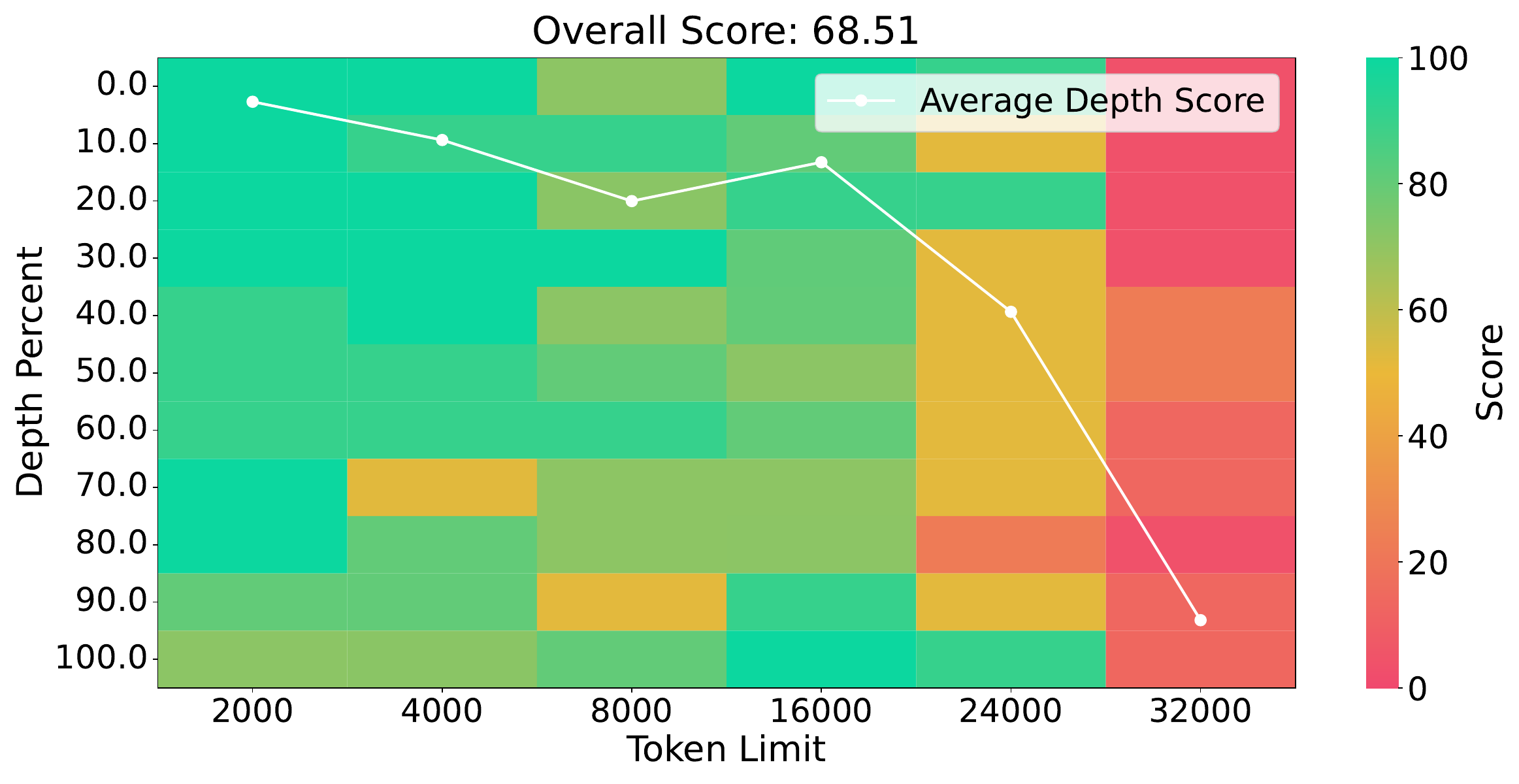}
        \caption{LLaDA-1.5 with $\lambda=31$}
        \label{llada_1_5_niah_lambda31}
    \end{subfigure}
    \caption{NIAH Results of LLaDA-1.5~\citep{zhu2025llada} with different $\lambda$.\label{llada_1_5_niah_ntk}}
\end{minipage}
\end{figure}

\begin{figure}[!tb]
\begin{minipage}{0.98\textwidth}
    \begin{subfigure}[b]{0.48\linewidth}
        \centering
        \includegraphics[width=\linewidth]{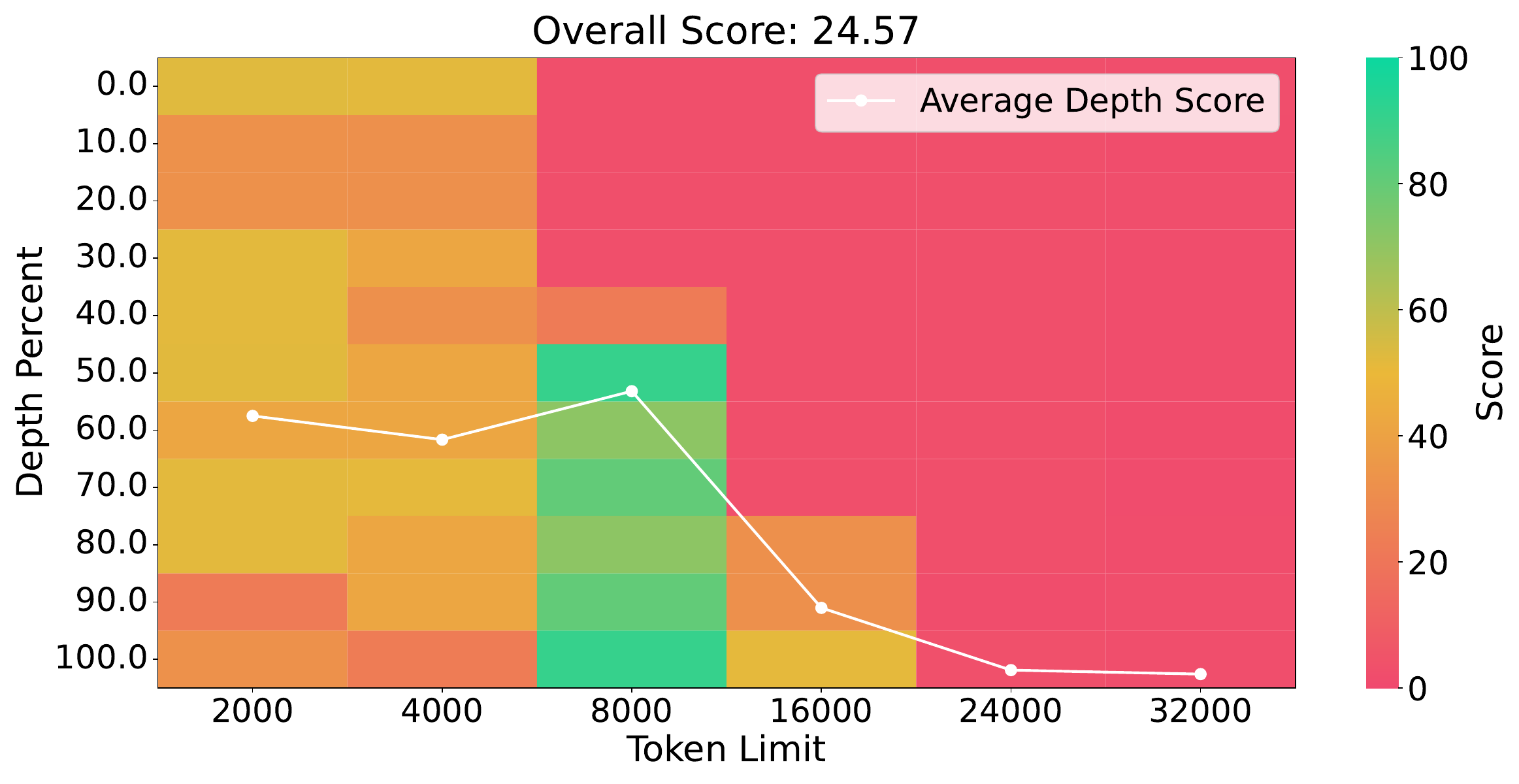}
        \caption{LLaDA-8B-Instruct with $s=1$}
        \label{llada_8b_chat_niah_step1}
    \end{subfigure}
    \hfill
    \begin{subfigure}[b]{0.48\linewidth}
        \centering
        \includegraphics[width=\linewidth]{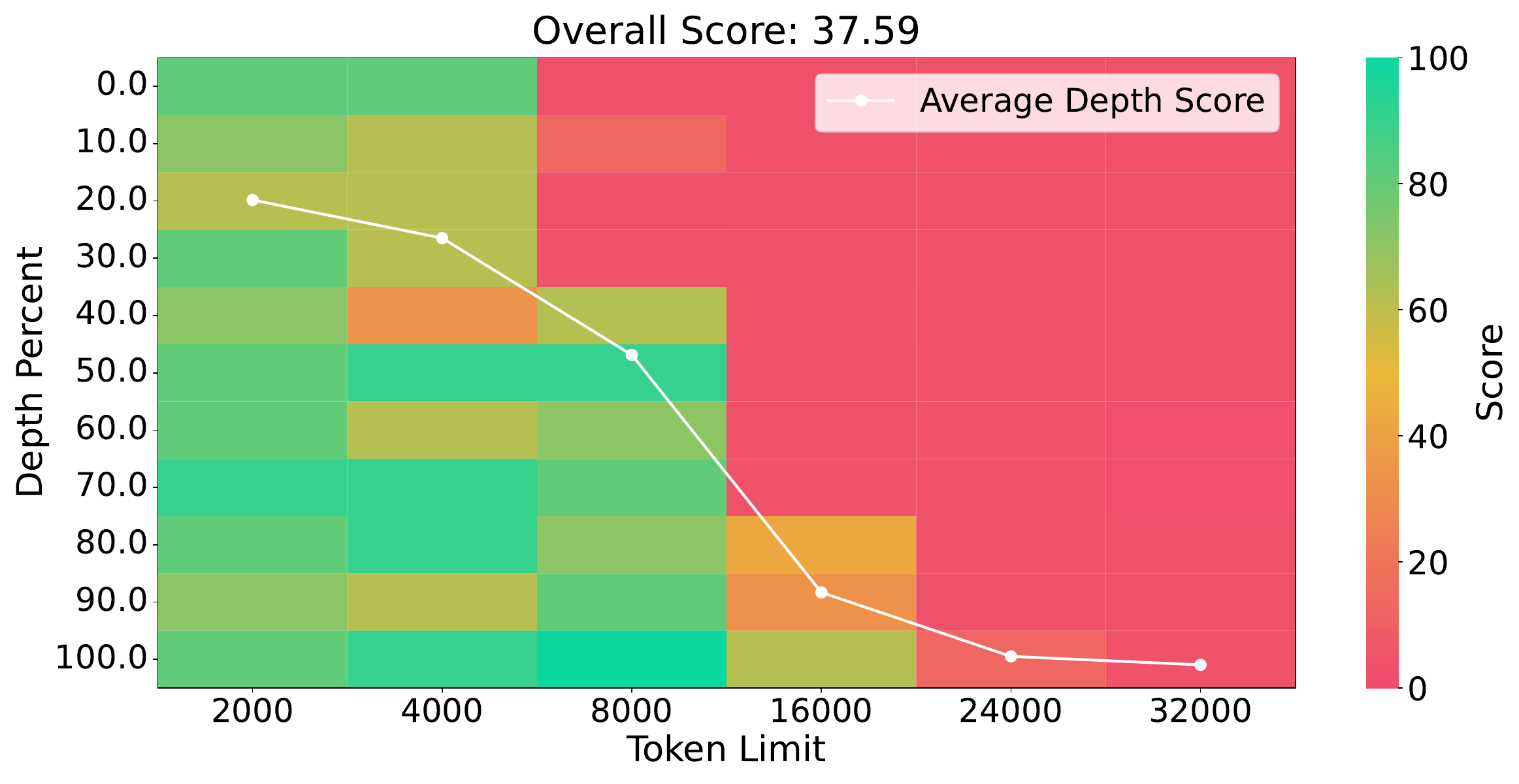}
        \caption{LLaDA-8B-Instruct with $s=4$}
        \label{llada_8b_chat_niah_step4}
    \end{subfigure}
    \vskip\baselineskip
    \begin{subfigure}[b]{0.48\linewidth}
        \centering
        \includegraphics[width=\linewidth]{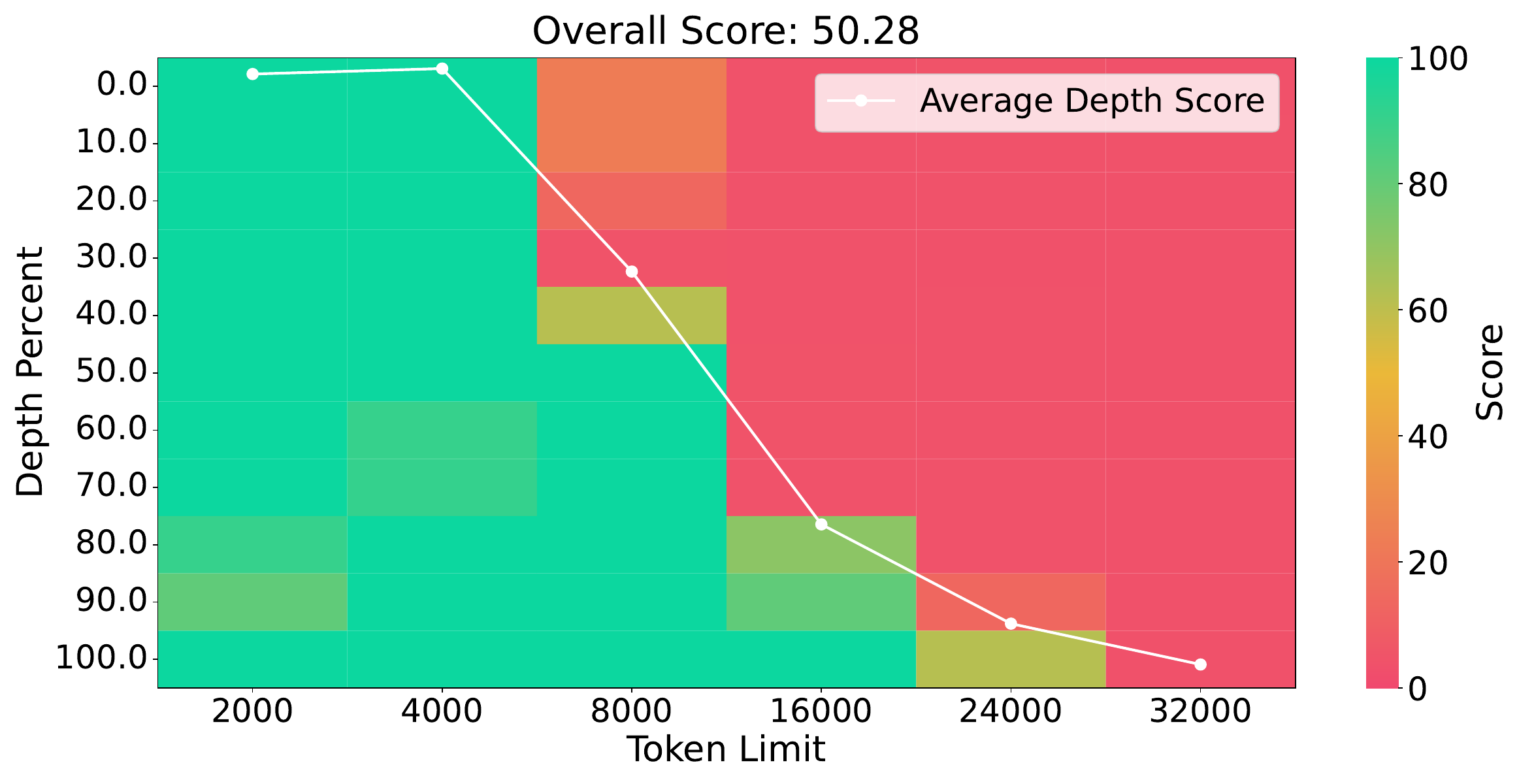}
        \caption{LLaDA-8B-Instruct with $s=8$}
        \label{llada_8b_chat_niah_step8}
    \end{subfigure}
    \hfill
    \begin{subfigure}[b]{0.48\linewidth}
        \centering
        \includegraphics[width=\linewidth]{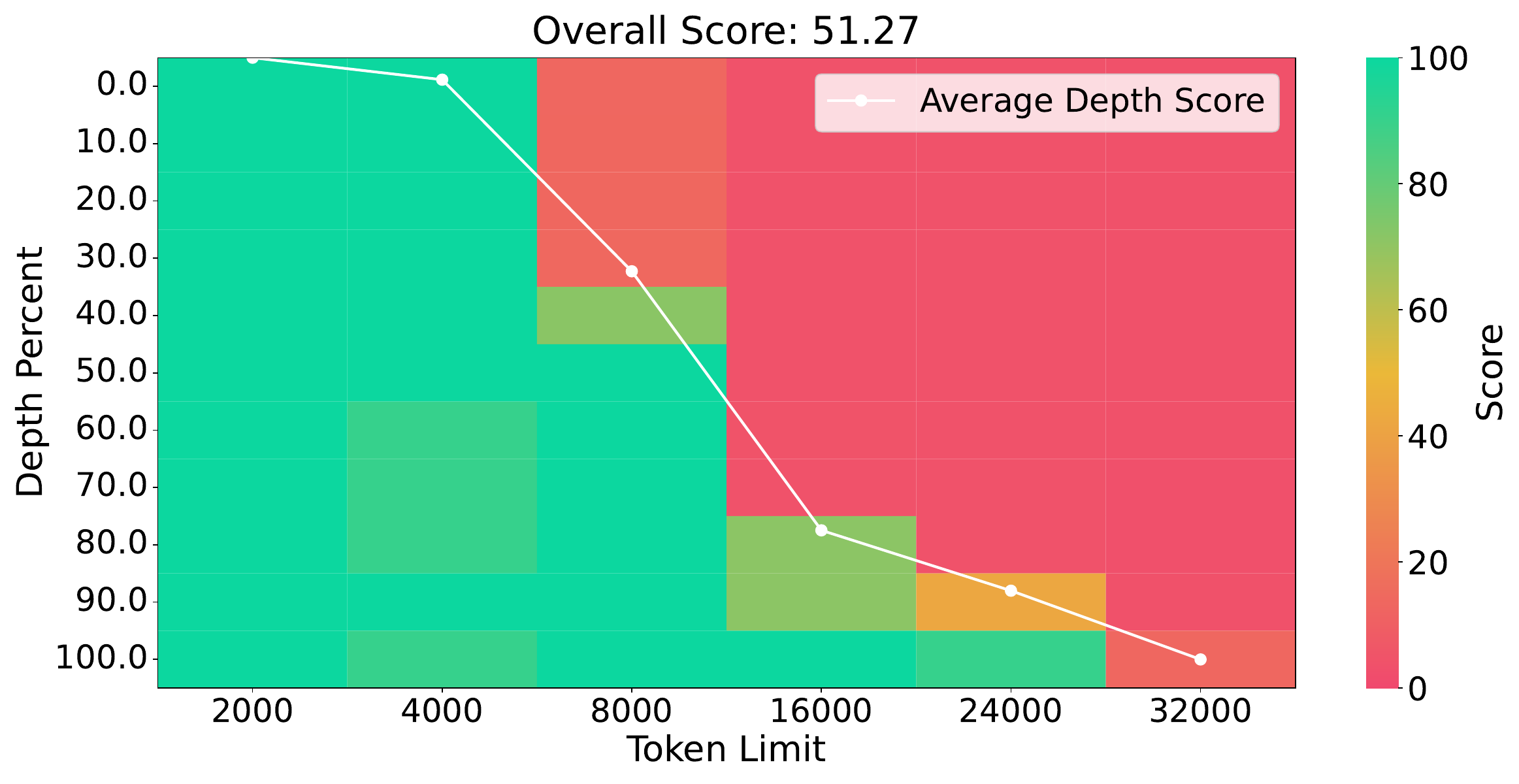}
        \caption{LLaDA-8B-Instruct with $s=16$}
        \label{llada_8b_chat_niah_step16}
    \end{subfigure}
    \caption{NIAH Results of LLaDA-8B-Instruct~\citep{nie2025large} with different sampling steps.\label{llada_chat_step}}
\end{minipage}
\end{figure}

\begin{figure}[!tb]
\begin{minipage}{0.98\textwidth}
    \begin{subfigure}[b]{0.48\linewidth}
        \centering
        \includegraphics[width=\linewidth]{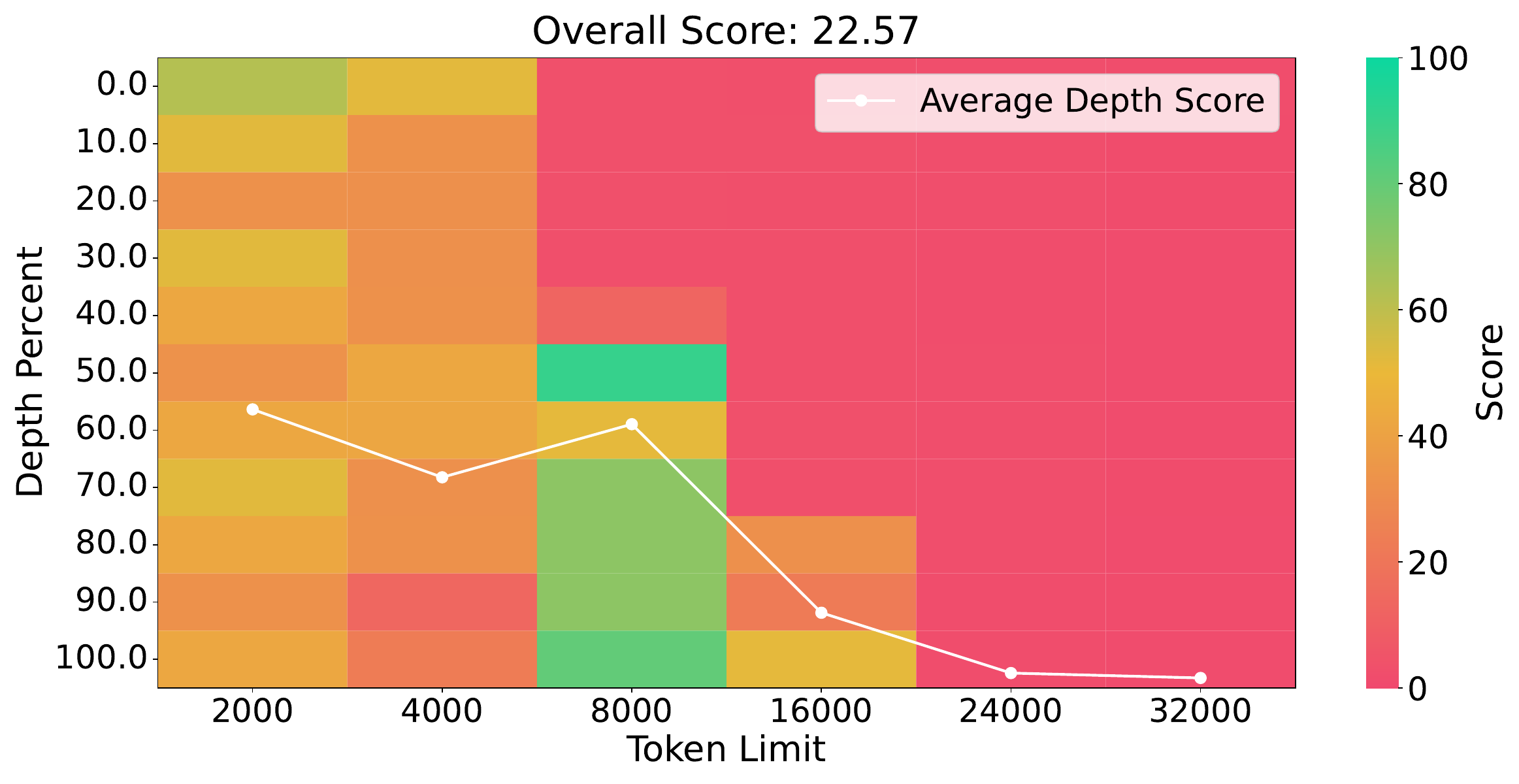}
        \caption{LLaDA-1.5 with $s=1$}
        \label{llada_1_5_niah_step1}
    \end{subfigure}
    \hfill
    \begin{subfigure}[b]{0.48\linewidth}
        \centering
        \includegraphics[width=\linewidth]{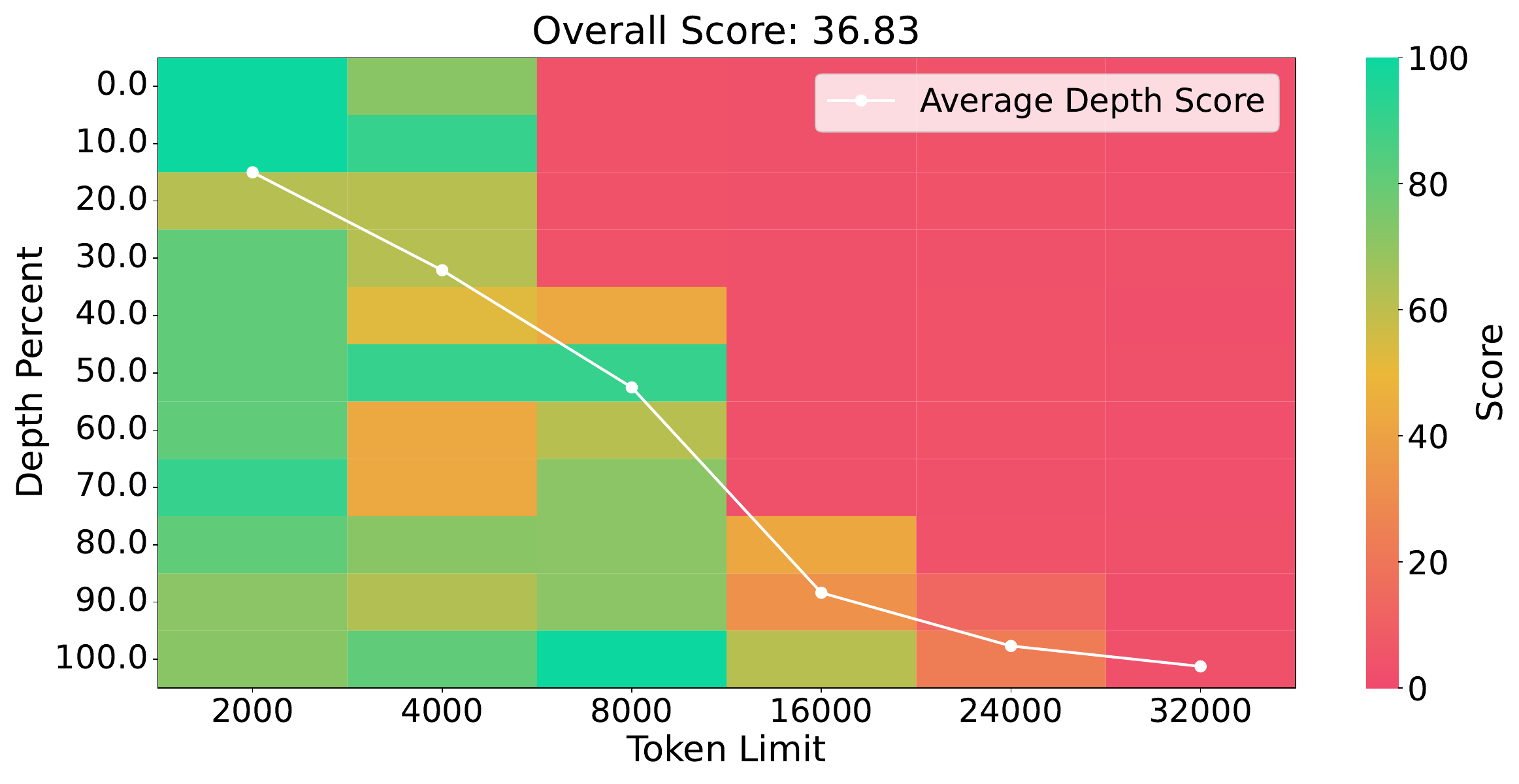}
        \caption{LLaDA-1.5 with $s=4$}
        \label{llada_1_5_niah_step4}
    \end{subfigure}
    \vskip\baselineskip
    \begin{subfigure}[b]{0.48\linewidth}
        \centering
        \includegraphics[width=\linewidth]{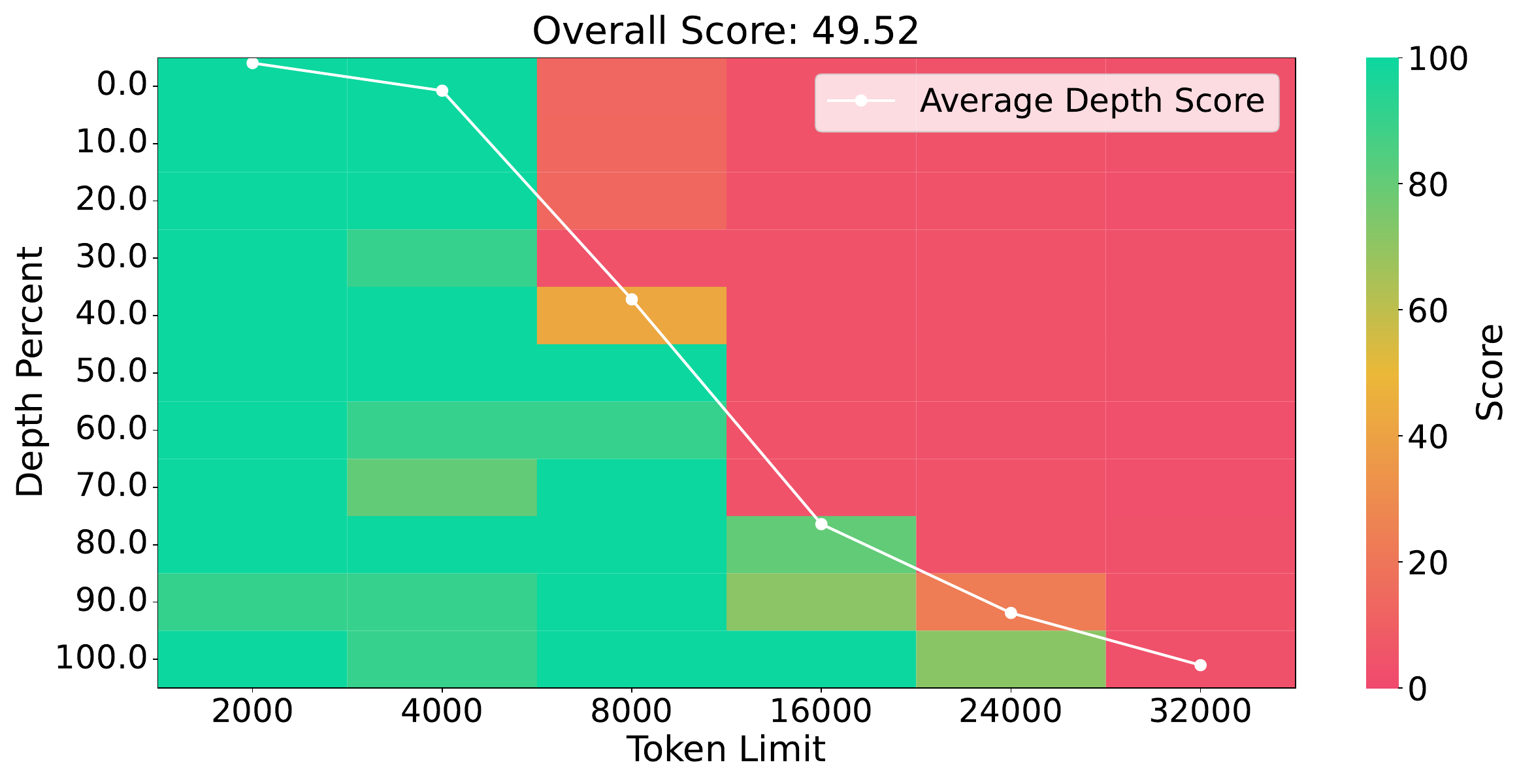}
        \caption{LLaDA-1.5 with $s=8$}
        \label{llada_1_5_niah_step8}
    \end{subfigure}
    \hfill
    \begin{subfigure}[b]{0.48\linewidth}
        \centering
        \includegraphics[width=\linewidth]{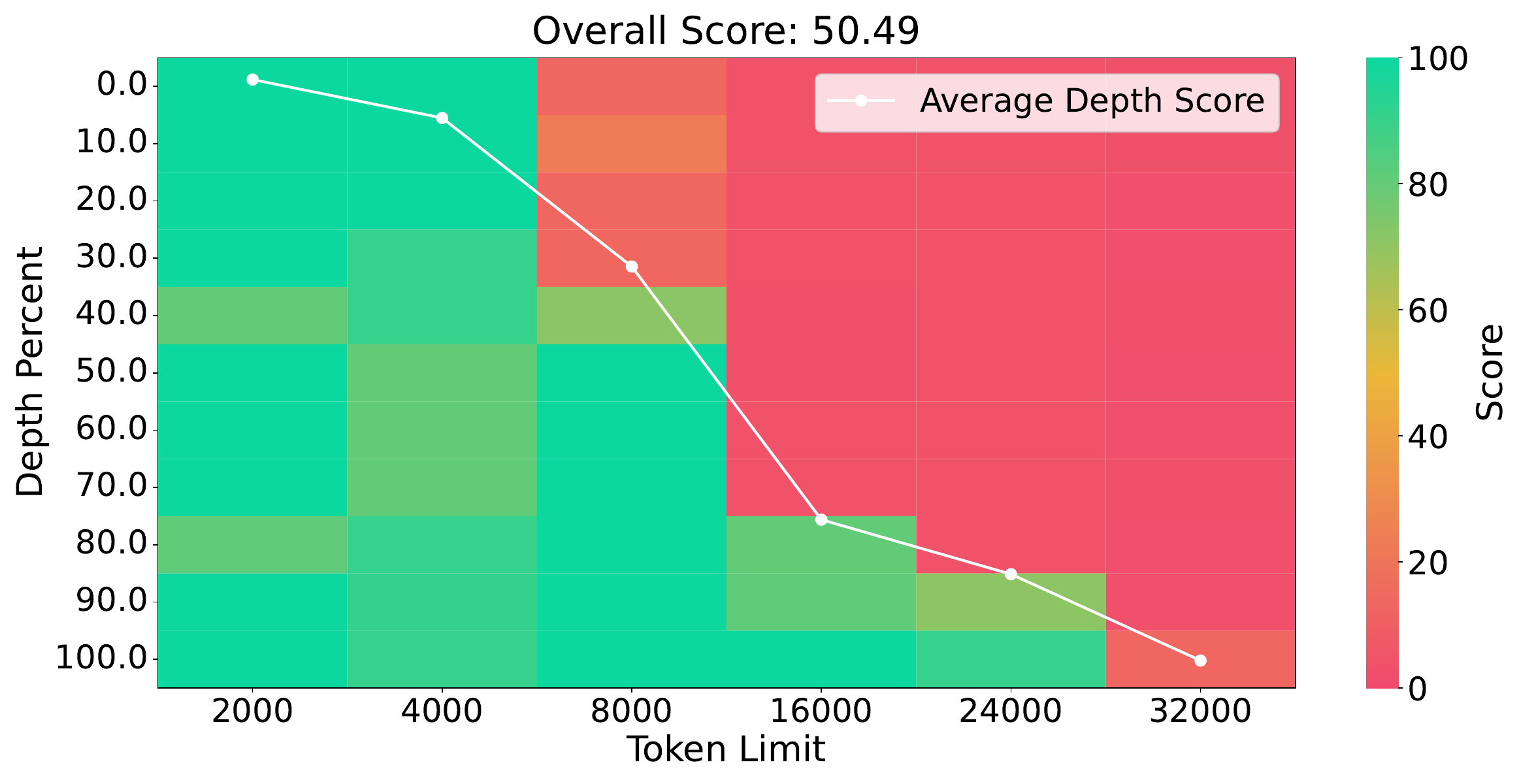}
        \caption{LLaDA-1.5 with $s=16$}
        \label{llada_1_5_niah_step16}
    \end{subfigure}
    \caption{NIAH Results of LLaDA-1.5~\citep{zhu2025llada} with different sampling steps.\label{llada_1_5_step}}
\end{minipage}
\end{figure}

We report the NIAH results of LLaDA-8B-Instruct~\citep{nie2025large} and LLaDA~\citep{zhu2025llada} with different sampling steps $s$ in Figure~\ref{llada_chat_step} and Figure~\ref{llada_1_5_step} respectively, similar to Figure~\ref{llada_base_step}. 
In Figure~\ref{llada_1_5_niah_ntk}, we report the NIAH results of pre-trained and NTK-scaled LLaDA-1.5~\citep{zhu2025llada}. LLaDA-1.5 still supports 4k context length and has a local perception in direct length extrapolation. We use the same scaling factors as LLaDA and achieve similar length extrapolation effects consistent with the prediction of the scaling law of RoPE-based length extrapolation~\citep{liu2023scaling}.

\begin{figure}[!tb]
\begin{minipage}{0.98\textwidth}
    \begin{subfigure}[b]{0.48\linewidth}
        \centering
        \includegraphics[width=\linewidth]{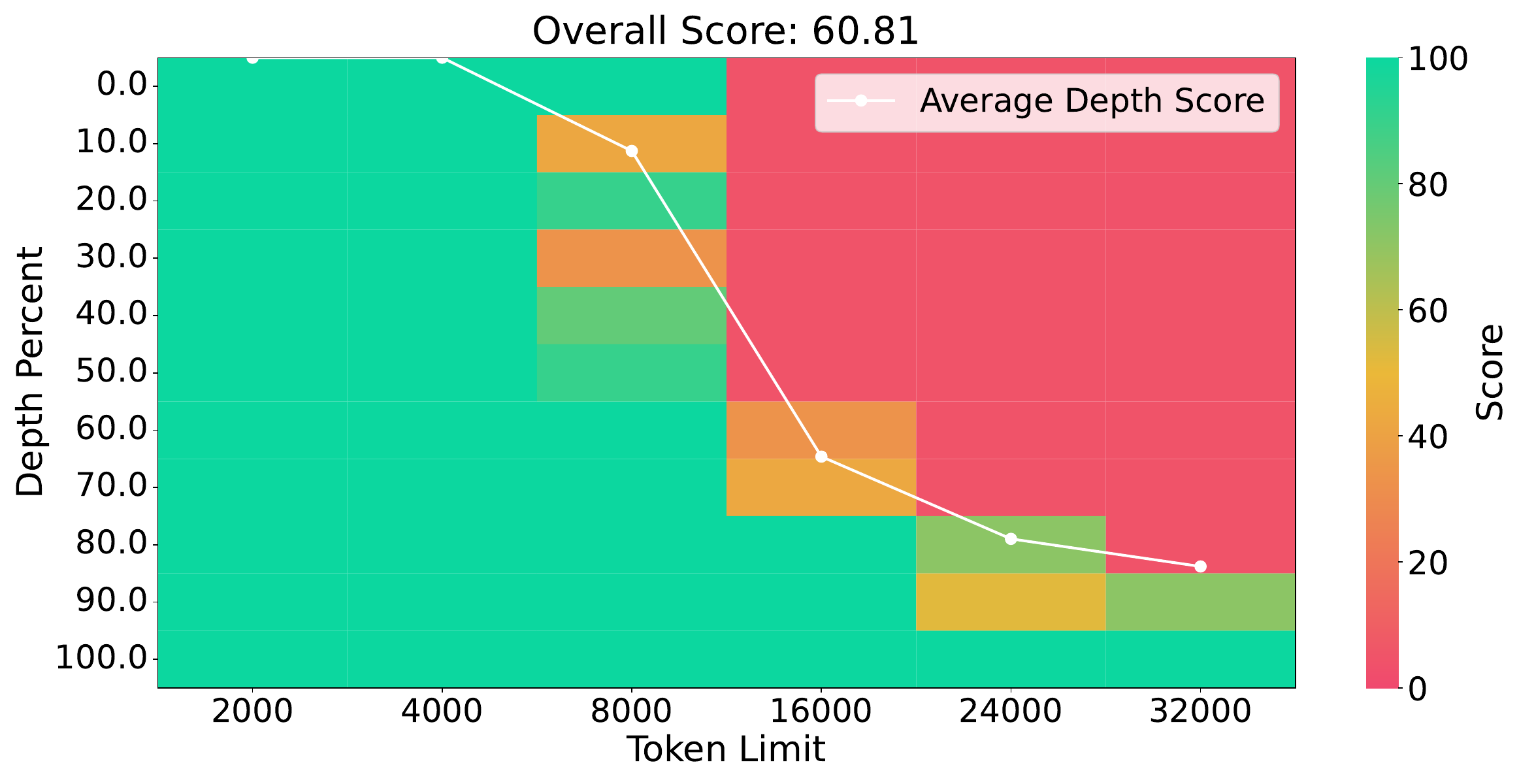}
        \caption{Pretrained Dream-v0-7B-Base}
        \label{dream_0_base_niah}
    \end{subfigure}
    \hfill
    \begin{subfigure}[b]{0.48\linewidth}
        \centering
        \includegraphics[width=\linewidth]{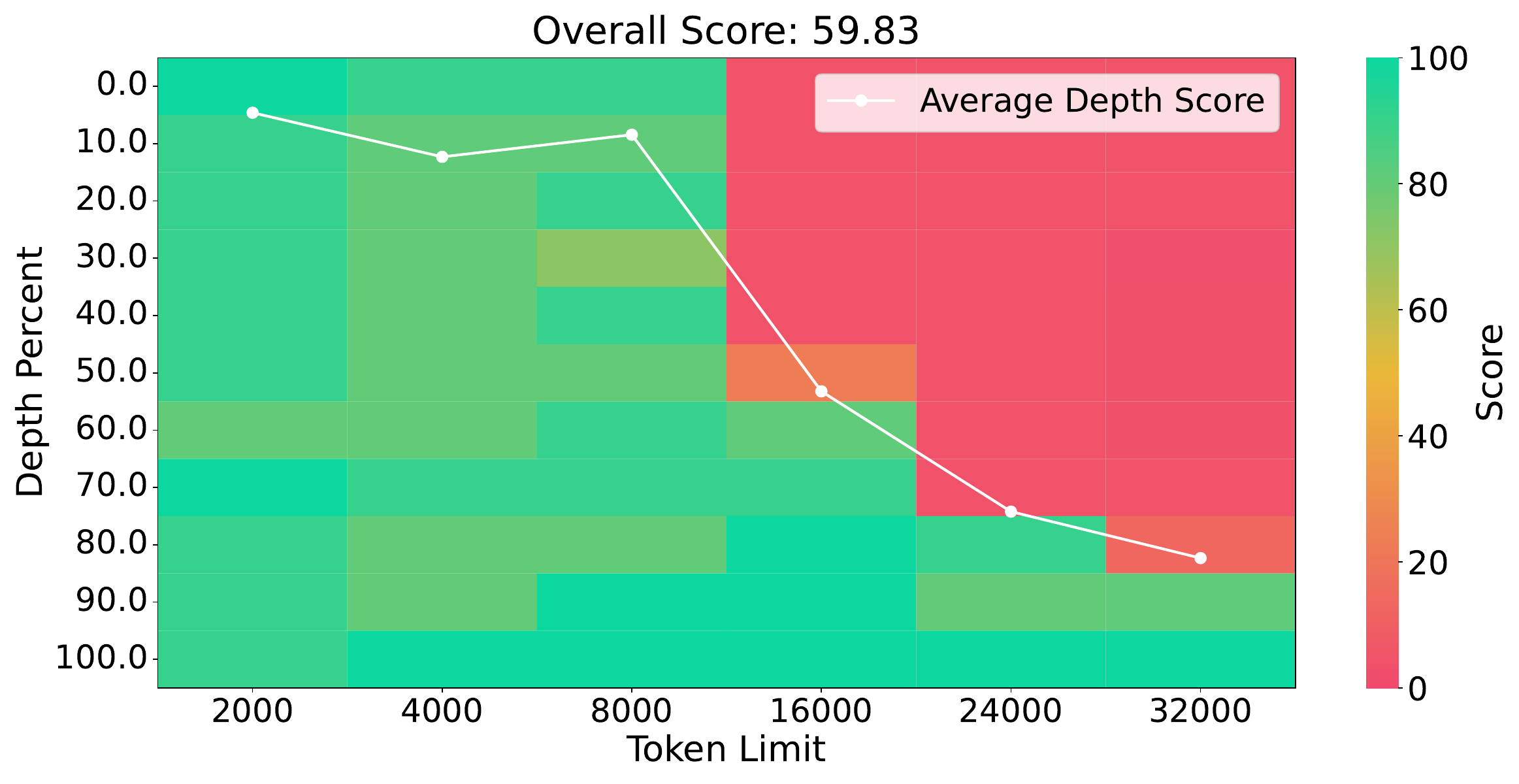}
        \caption{Pretrained Dream-v0-7B-Instruct}
        \label{dream_0_chat_niah}
    \end{subfigure}
    \caption{NIAH Results of Dream-v0-7B Series~\citep{dream2025}.\label{dream_0_niah}}
\end{minipage}
\end{figure}

In Figure~\ref{dream_0_niah}, we also report the NIAH performance of Dream-v0-7B-Base and Dream-v0-7B-Instruct. We find that both diffusion LLMs still have a local perception in direct extrapolation. Since Dream-v0 Series is trained not from scratch but initialized from an auto-regressive LLM, Qwen2.5-7B~\citep{dream2025}, it is hard to define the exact context length of pre-training. Therefore, we only compare the long-context performance of pretrained models to verify the task characteristics as shown in Table~\ref{tab_longbench_dream} and Table~\ref{tab_ruler_dream}. Dream Series achieves a comparable and even better performance in LongBench~\citep{bai2023longbench}, and demonstrates strength in synthetic QA and weakness in aggregation tasks in RULER~\citep{hsieh2024ruler} within a 4k context.

\begin{table}[!tb]
\tabcolsep=0.1cm
\centering\small
\begin{tabular}{lcccccccccccccc}
\toprule
& \multicolumn{7}{c}{\textbf{4k}} & \multicolumn{7}{c}{\textbf{8k}} \\
\cmidrule(lr){2-8} \cmidrule(lr){9-15} 
& SD & MD & Sum & ICL & Syn & Code & Avg & SD & MD & Sum & ICL & Syn & Code & Avg \\
\midrule
\textbf{\textit{Dream-v0-7B-Base}} & 22.8 & 20.7 & 34.1 & 36.3 & 64.4 & 68.3 & \underline{38.8} & 24.0 & 21.6 & 35.9 & 40.3 & 73.9 & 68.9 & 41.7 \\
\textbf{\textit{Dream-v0-7B-Instruct}} & 23.8 & 21.7 & \textbf{35.1} & 37.3 & \textbf{65.4} & \textbf{69.3} & \textbf{39.8} & 25.0 & 22.6 & \textbf{36.9} & 41.3 & \textbf{74.9} & \textbf{69.9} & \textbf{42.7} \\
\textbf{\textit{LLaMA3-8B-Base}} & 17.2 & 18.7 & 25.0 & \textbf{41.7} & 47.6 & 66.5 & 33.6 & 18.2 & 18.3 & 26.1 & \textbf{44.5} & 49.6 & 69.4 & 35.1 \\
\textbf{\textit{LLaMA3-8B-Instruct}} & \textbf{31.9} & \textbf{26.1} & 33.6 & 39.6 & 46.6 & 55.9 & 37.0 & \textbf{37.5} & \textbf{28.3} & 34.7 & 40.7 & 62.8 & 56.1 & \underline{41.9} \\
\bottomrule
\end{tabular}
\caption{Results of Dream-v0-7B~\citep{dream2025} and LLaMA3-8B~\citep{meta2024llama} on LongBench~\citep{bai2023longbench} under 4k and 8k context length.\label{tab_longbench_dream}}
\end{table}

\begin{table}[!tb]
\tabcolsep=0.12cm
\centering\small
\begin{tabular}{lcccccccccccc}
\toprule
& \multicolumn{4}{c}{\textbf{4k}} & \multicolumn{4}{c}{\textbf{8k}} & \multicolumn{4}{c}{\textbf{16k}} \\
\cmidrule(lr){2-5} \cmidrule(lr){6-9} \cmidrule(lr){10-13} 
~ & NIAH & AGG & QA & Avg & NIAH & AGG & QA & Avg & NIAH & AGG & QA & Avg \\ 
\midrule
\textbf{\textit{Dream-v0-7B-Base}} & 90.6 & 68.3 & \textbf{78.5} & 83.6 & 68.1 & 49.7 & 61.5 & 62.9 & 24.3 & 26.5 & 45.0 & 28.0 \\
\textbf{\textit{Dream-v0-7B-Instruct}} & 95.6 & 74.6 & 76.5 & 87.8 & 72.3 & 55.8 & 58.0 & 66.3 & \textbf{31.0} & \textbf{29.5} & \textbf{46.0} & \textbf{33.0} \\
\textbf{\textit{LLaMA3-8B-Base}} & \textbf{99.8} & \textbf{98.1} & 67.5 & \textbf{94.4} & 99.6 & \textbf{93.5} & \textbf{63.0} & \textbf{92.5} & 0.0 & 0.0 & 0.0 & 0.0 \\
\textbf{\textit{LLaMA3-8B-Instruct}} & 99.6 & 97.2 & 68.5 & 94.3 & \textbf{98.2} & 92.6 & 54.0 & 90.1 & 0.0 & 0.0 & 0.0 & 0.0 \\
\bottomrule
\end{tabular}
\caption{Results of Dream-v0-7B~\citep{dream2025} and LLaMA3-8B~\citep{meta2024llama} on RULER~\citep{hsieh2024ruler} under 4k, 8k and 16k context length.\label{tab_ruler_dream}}
\end{table}

\end{document}